\documentclass[10pt,twocolumn,letterpaper]{article}

\usepackage[pagenumbers]{cvpr} 

\usepackage{multirow}
\usepackage{xcolor}
\usepackage{multicol}
\usepackage{pifont}
\usepackage{amsmath} 
\usepackage{algorithm,algorithmic}
\usepackage{xcolor}
\usepackage{listings}
\makeatletter
\AfterEndEnvironment{algorithm}{\let\@algcomment\relax}
\AtEndEnvironment{algorithm}{\kern2pt\hrule\relax\vskip3pt\@algcomment}
\let\@algcomment\relax
\newcommand\algcomment[1]{\def\@algcomment{\footnotesize#1}}
\renewcommand\fs@ruled{\def\@fs@cfont{\bfseries}\let\@fs@capt\floatc@ruled
  \def\@fs@pre{\hrule height.8pt depth0pt \kern2pt}%
  \def\@fs@post{}%
  \def\@fs@mid{\kern2pt\hrule\kern2pt}%
  \let\@fs@iftopcapt\iftrue}
\makeatother

\definecolor{cvprblue}{rgb}{0.21,0.49,0.74}
\usepackage[pagebackref,breaklinks,colorlinks,citecolor=cvprblue]{hyperref}

\def\paperID{*****} 
\def\confName{CVPR}
\def\confYear{2024}

\title{
\raisebox{-1.1ex}{\protect\includegraphics[height=1.1cm]{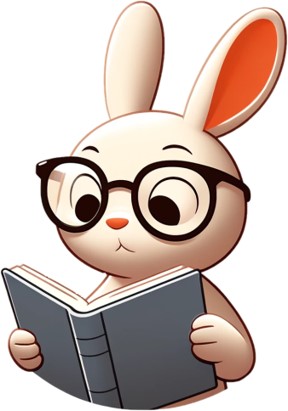}}
Self-Supervised Visual Preference Alignment}

\author{
	Ke Zhu$^{1,2}$ \quad
	Liang Zhao$^{4}$ \quad
	Zheng Ge$^{3,4}$ \quad
	Xiangyu Zhang$^{3,4}$ \\
$^1$State Key Laboratory for Novel Software Technology, Nanjing University, China \\
$^2$School of Artificial Intelligence, Nanjing University, China \\
$^3$MEGVII Technology \\
$^4$StepFun Intelligent Technology \\
{\tt\small zhuk@lamda.nju.edu.cn, \{zhaoliang06, gezheng, zhangxiangyu\}@megvii.com}
}

\begin{document}
\maketitle

\begin{abstract}
    This paper makes the first attempt towards unsupervised preference alignment in Vision-Language Models (VLMs). We generate chosen and rejected responses with regard to the original and augmented image pairs, and conduct preference alignment with direct preference optimization. It is based on a core idea: properly designed augmentation to the image input will induce VLM to generate false but hard negative responses, which helps the model to learn from and produce more robust and powerful answers. The whole pipeline no longer hinges on supervision from GPT-4 or human involvement during alignment, and is highly efficient with few lines of code. With only 8k randomly sampled unsupervised data, it achieves 90\% relative score to GPT-4 on complex reasoning in LLaVA-Bench, and improves LLaVA-7B/13B by 6.7\%/5.6\% score on complex multi-modal benchmark MM-Vet. Visualizations shows its improved ability to align with user-intentions. A series of ablations are firmly conducted to reveal the latent mechanism of the approach, which also indicates its potential towards further scaling. Code are available in \textcolor{cvprblue}{https://github.com/Kevinz-code/SeVa}.
\end{abstract}

\section{Introduction}
Large Vision-Language Models have recently emerged and greatly advanced current researches~\cite{LLM_Flamingo,LLM_Llava,ACM_Retrieval,ACM_VideoRetrieval}. Representative methods in this family, such as MiniGPT4~\cite{LLM_MiniGPT4}, LLaVA~\cite{LLM_Llava} and InstructBLIP~\cite{InstructBLIP}, try to properly align vision signals to Large Language Models (LLMs) to better conduct multi-modal comprehension. These methods usually undergo a pretraining stage with abundant image-text pairs for modality alignment before supervised finetuned (SFT) with 
academic~\cite{InstructBLIP} or GPT-4 generated~\cite{ShareGPT4V,Scale_SFT} instruction following data. There are variants of them (e.g., LLaVA-based) that try to improve the VLMs' instruction following ability by leveraging high quality pretraining pairs~\cite{ShareGPT4V,VILA} or scaling up SFT database~\cite{DiffusionSFT,Scale_SFT,LVIS-Instruct4V}.

Despite their success~\cite{ShareGPT4V} in boosting the comprehension skills of current VLMs, these models are not properly aligned with user-intentions. As a result, they lack the ability to reject samples and could induce more unintended output. For instance, Instruct4V~\cite{LVIS-Instruct4V} construct 220k SFT data from GPT-4, but still struggle on hallucination benchmarks POPE~\cite{POPE}. The same can be observed in Fig.~\ref{fig:question-answer}, where LLaVA failed to follow user instructions and provide meaningless information in its answers. Recently, there \emph{are} few trials that incorporate alignment techniques into vision-language fields~\cite{HA-DPO,VLLM_DPO,LLaVa-RLHF}. However, we found that they only emphasize on quite limited task domains~\cite{HA-DPO}, and, more importantly, \emph{their data construction pipeline requires extra knowledge source either from GPT-4 or human feedbacks}. This might hinder them from further data scaling, as preference data is \emph{not cheap} (e.g., a 10k collected human-evaluated instances requires a cost of 3000\$~\cite{LLaMA-VID}).


\begin{figure}
	\centering
    \includegraphics[width=0.95\linewidth]{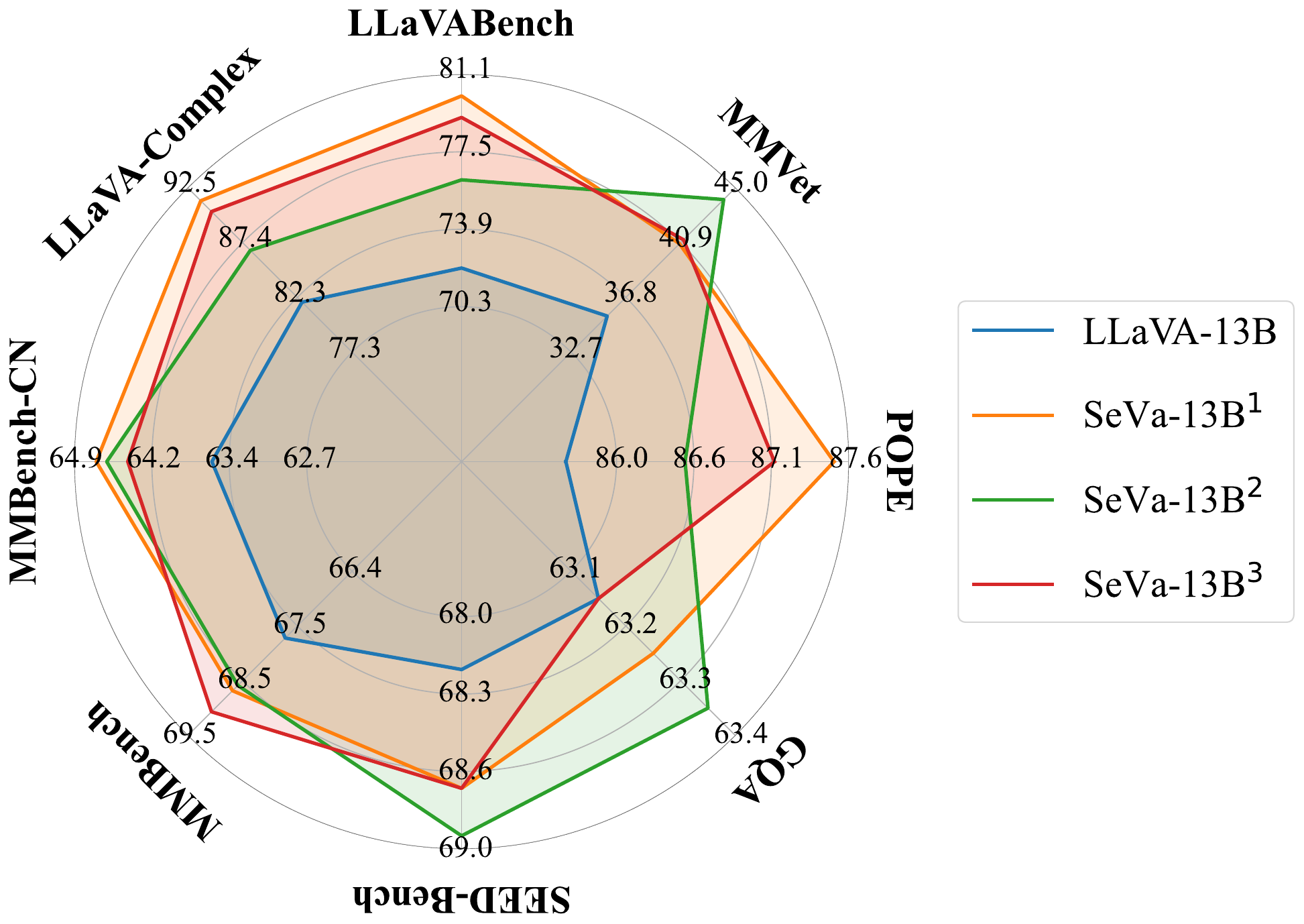}
	\caption{Illustration of the baseline LLaVA-13B (v1.5) and the proposed SeVa-13B. Here we demonstrate three variants of SeVa with different sampled seed to obtain the un-labeled dataset (the image-text pairs used for DPO sample generation, \cf Alg.~\ref{alg:code}).}
	\label{fig:seva-figure1}
\end{figure}

In this paper, we propose a self-supervised paradigm that can \emph{freely}  obtain arbitrary number of preference samples  with minimal code implementation (c.f., Alg.~\ref{alg:code}). Simply but, we have enjoyed the advantage of this pipeline and observed a significant improvement in the capabilities of current VLMs from various aspects: stronger chain-of-thought skills, better OCR~\cite{OCRVQA} ability, proper alignment with user-intentions, less hallucinations, etc (\cf Fig.~\ref{fig:question-answer}).

\begin{figure*}
	\centering
	\begin{subfigure}{0.225\linewidth}
		\includegraphics[width=1.0\linewidth]{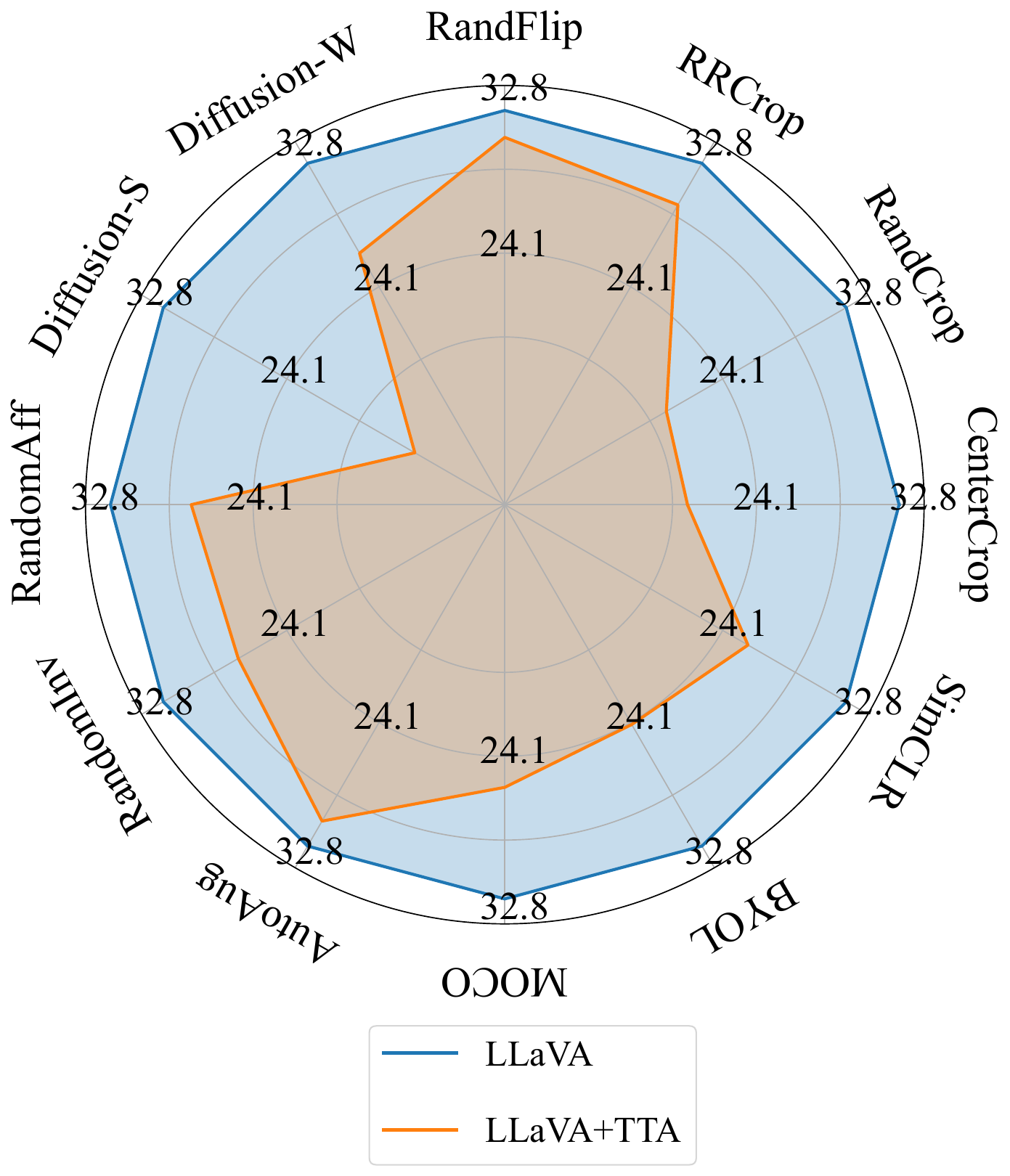}
    	\caption{MMVet}
		\label{fig:motivation:mmvet}
    \end{subfigure} 
    \hspace{15pt}
    \begin{subfigure}{0.225\linewidth}
		\includegraphics[width=1.0\linewidth]{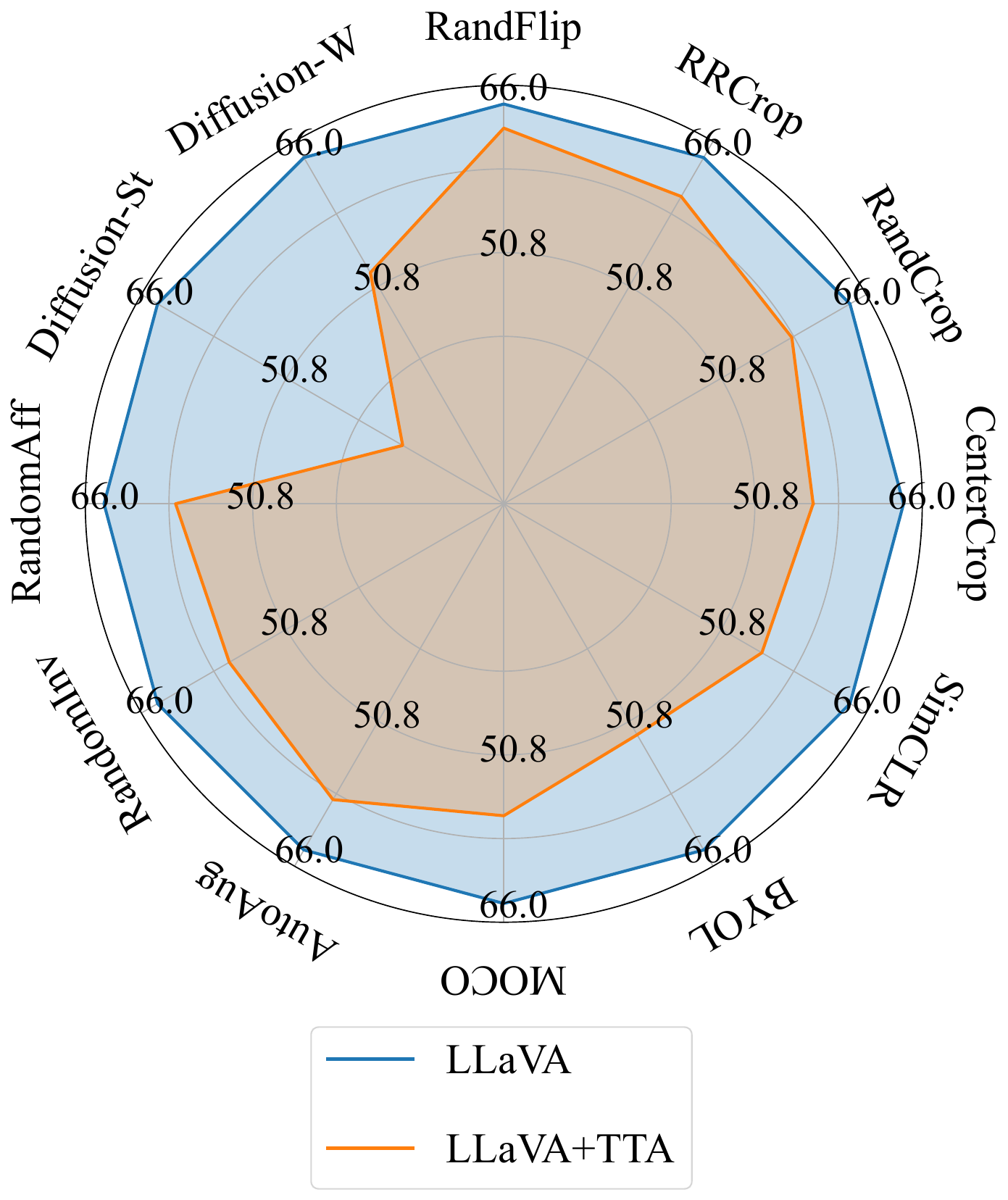}
    	\caption{MMBench}
		\label{fig:motivation:mmbench}
    \end{subfigure}
    \hspace{15pt}
    \begin{subfigure}{0.225\linewidth}
		\includegraphics[width=1.0\linewidth]{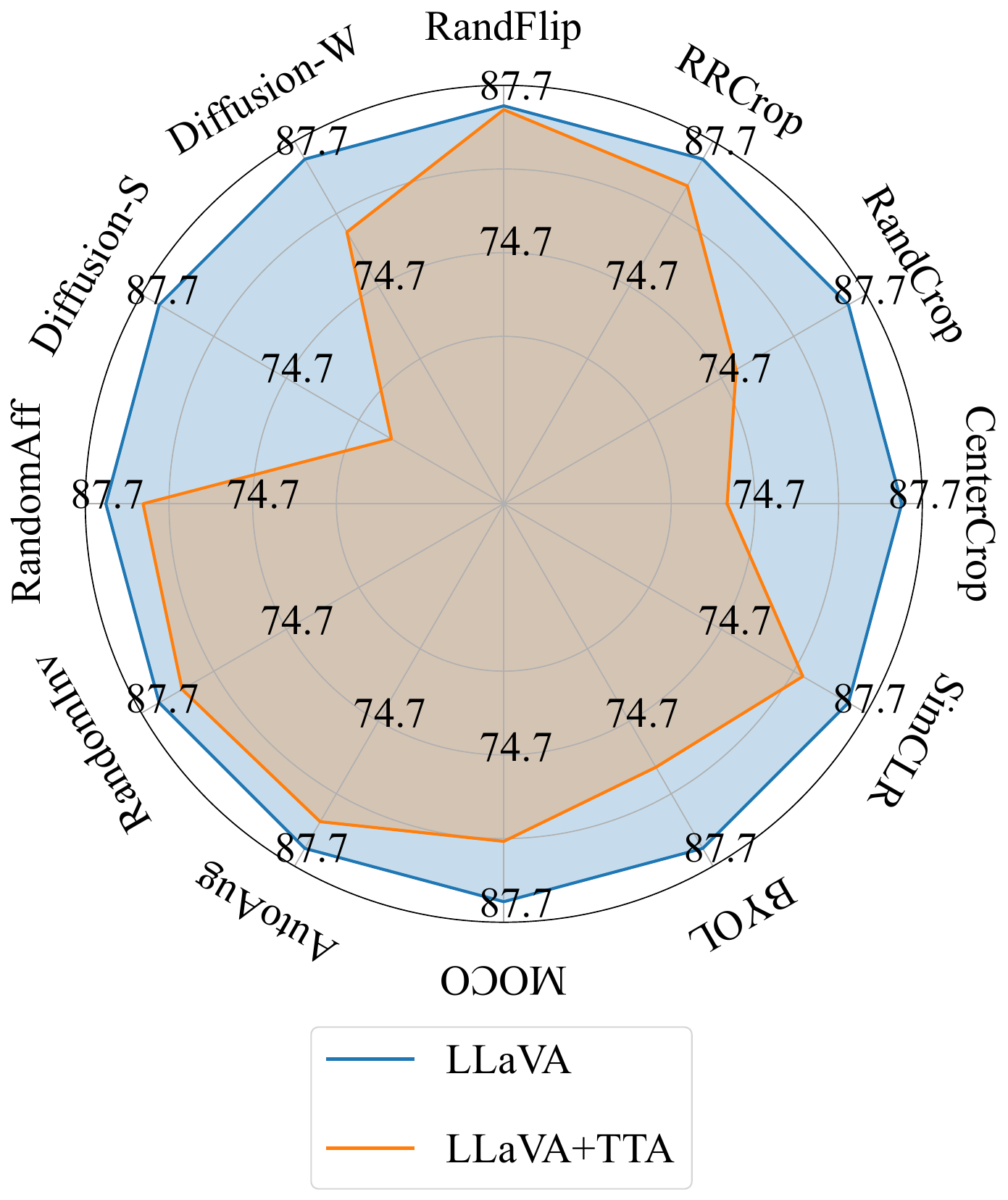}
		\caption{POPE}
		\label{fig:motivation:pope}
	\end{subfigure}

	\caption{The test-time image augmentations (TTA) plugged into LLaVA-1.5 on three benchmarks. We involve standard augmentation: RandFlip, RandomResizedCrop (`RRCrop'), RandomCrop, CenterCrop, RandomAffine, RandomInvert and AutoAug; diffusion noise augmentation: Diffusion-Weak (`W') and Diffusion-Strong ('S'); mixtures: strategies adopted in MOCO~\cite{MOCO}, BYOL~\cite{BYOL} and SimCLR~\cite{SimCLR}.}
	\label{fig:data-aug-tta}

\end{figure*}

Our motivation came from an interesting discovery in Fig.~\ref{fig:data-aug-tta}, where we apply commonly adopted image augmentations in \emph{visual contrastive learning}~\cite{MOCO,MLS,Cropping} during LLaVA inference. The effect of all these test-time augmentations are evaluated on three common multi-modal benchmarks (MMVet~\cite{MMVet}, POPE~\cite{POPE}, MMBench~\cite{MMBench}). It is not surprising to observe in Fig.~\ref{fig:data-aug-tta} that \emph{vision-language models are quite sensitive to data-augmentations} and, slightly distortion will induce the model to output different semantic tokens. Then, an natural question arises: \emph{are the original and distorted responses valuable to construct preference data for DPO training?} Since this will totally free the data construction from \emph{any} extra source, and might relieve the difficulty of scaling up preference data as well~\cite{zhaoAlign}.

\begin{algorithm}[t]
\caption{Pseudocode of SeVa in a PyTorch style.}
\label{alg:code}

\definecolor{codeblue}{rgb}{0.25,0.5,0.5}
\lstset{
  backgroundcolor=\color{white},
  basicstyle=\fontsize{7.2pt}{7.2pt}\ttfamily\selectfont,
  columns=fullflexible,
  breaklines=true,
  captionpos=b,
  commentstyle=\fontsize{7.2pt}{7.2pt}\color{codeblue},
  keywordstyle=\fontsize{7.2pt}{7.2pt},
}
\begin{lstlisting}[language=python, mathescape=true,morekeywords={yl',yw'}]
# Q, I: question set, image set
# model: the SFT vision-language model
# C, R:  chosen and rejected answer set
# T: sampled data augmentation

# randomly sample data pair, generate answer 
Q, I = random.sample(data)
C, R = model(I, Q), model(T(I), Q)

# filtering equal answers
idx = (C != R)
C$_t$, R$_t$ = C[idx], R[idx]
Q$_t$, I$_t$ = Q[idx], I[idx]

# DPO training, omit reference model for simplicity
DPOTrain(model, (Q$_t$, I$_t$), (C$_t$, R$_t$))

\end{lstlisting}
\label{algo:1}
\end{algorithm}

Our motivation was then firmly verified by quantitative experiments in Table~\ref{tab:data-aug-improve}, in which we randomly sampled 8k image-question pairs from the subset of LLaVA665k~\cite{LLaVa1.5} (\cf Sec.~\ref{sec:exp-settings} for details), and choose 5 augmentations to generate the preference data using LLaVA-7B. All these self-generated data are again fed into the \emph{same} LLaVA-7B model for DPO training. As Table~\ref{tab:data-aug-improve} shows, all the augmentations are helpful for model comprehension, especially on GPT-4 evaluated benchmark MM-Vet, which makes our motivation valid. In the meanwhile, we found that either too strong (e.g., Diffusion-S) or too weak data augmentation (e.g., RandFlip) are sub-optimal for the whole pipeline, and medium is the best (e.g., Diffusion-W, MOCO). Our final conjecture is: \emph{self-generated augmentation pairs are suitable preference sample to improve multi-modal abilities, and hard negatives are most preferred.}

We name our method: \textbf{Se}lf-supervised \textbf{V}isual preference \textbf{a}lignment (SeVa), and summarize its whole pipeline in Alg.~\ref{algo:1}. Though being embarassingly simple with few lines of code to implement, we mathematically showcase its strong relation with visual contrastive learning in maxmizing a preference distribution, where SeVa could probably be viewed as a special form of contrastive learning with one negative sample. This makes SeVa easily extendable if more negatives are involved (\cf appendix for details).

Experiments in common multi-modal benchmarks demonstrate the effectiveness of our methods, where SeVa improves the VLM's comprehension ability by a large margin (e.g., SeVa-7B even surpasses LLaVA-1.5-13B by 1.8\% in MMVet). In addition, we carefully design detailed ablations to reveal the latent mechanism of SeVa from different angles. We found that SeVa shows surprisingly results like detailed descriptions, less hallucinations, stronger OCR skills and chain-of-thought ability, etc. 

In summary, our contributions are:
\begin{itemize}
    \item For the first time, we conduct visual preference alignment in an un-supervised manner. The whole pipeline \emph{do not} require any GPT-4 or costly human annotated data.
    \item We found such an alignment pipeline have numerous benefits such as enhanced multi-modal comprehension ability, better alignment with user-intentions, etc. Visualizations and ablations firmly verify our hypothesis as well.
    \item Our method, SeVa, enjoys efficiency in pipeline and simplicity in implementation, which paves way for future preference alignment in visual-language domain.
\end{itemize}

\begin{table*}
	\small
	\centering
	\begin{tabular}{llllllllllllllll}
		\toprule[1pt]
		   \multicolumn{1}{c|}{\multirow{2}{*}{DPO data}}  & \multicolumn{7}{c|}{MMVet}& \multicolumn{2}{c|}{MMBench} & \multicolumn{4}{c}{POPE}\\

      & \multicolumn{1}{|c}{All} & rec  & ocr &know & gen & spat &  \multicolumn{1}{c|}{math} & en & \multicolumn{1}{c|}{cn} & All & rand & pop & \multicolumn{1}{c}{adv}\\
		\midrule[1pt]
        \textcolor{lightgray}{N/A} &  \textcolor{lightgray}{30.5} &  \textcolor{lightgray}{35.7}&  \textcolor{lightgray}{21.9} &  \textcolor{lightgray}{17.4} &  \textcolor{lightgray}{19.7} &  \textcolor{lightgray}{24.7} &  \textcolor{lightgray}{7.7} &  \textcolor{lightgray}{64.3} &  \textcolor{lightgray}{58.3} &  \textcolor{lightgray}{85.9} &  \textcolor{lightgray}{89.5} &  \textcolor{lightgray}{86.7} &  \textcolor{lightgray}{81.7}\\
        RandFlip & 33.7 & 37.2 & 26.4 & 21.8 & 23.9 & 29.1 & 7.7 & 64.4 & 58.3 & 86.7 & 89.2 & 87.1 & 83.9 \\
        RRCrop & 33.8 & 39.2 & 23.2 & 21.9 & 24.5 & 27.7 & 7.7 & 65.5 & 59.5 & 86.8 & 89.4 & 87.4 & 83.6\\
        AutoAug & 35.0 & 38.5 & 26.8 & 22.9 & 24.4 & 29.3 & 15.0 & 65.4 & 59.6 & 86.5 & 88.8 & 86.8 & 83.9\\
        Diffusion-W & 37.2 & 40.2 & 29.9 & 21.8 & 23.9 & 34.3 & 11.2 & 65.6 & 59.2 & 86.7 & 89.4 & 87.1 & 83.6 \\
        MOCO & 37.0 & 40.9 & 29.0 & 21.3 & 25.6 & 32.5 & 15.0 & 65.2 & 59.8& 86.6 & 89.1 & 87.1 & 83.7\\
        Diffusion-S & 34.6 & 38.8 & 26.5 & 20.5 & 23.4 & 32.0 & 11.5 & 65.2 & 58.2 & 86.6 & 89.2 & 87.5 & 83.3\\

		\bottomrule[1pt]
	\end{tabular}
\caption{Six data augmentations selected to generate preference data for DPO training (w/ LLaVA-1.5). Each strategy was adopted to distort the image (together with the questions) to produce rejected response, which is paired with the chosen response from the original image input (\cf Alg.~\ref{algo:1}). The models after DPO training with these preference pairs are then evaluated on MMVet~\cite{MMVet}, MMBench~\cite{MMBench} (in english `en' and chinese `cn') and POPE~\cite{POPE} benchmarks, respectively. Please refer to Sec.~\ref{sec:method:seva} and visualizations in appendix.}
\label{tab:data-aug-improve}
\end{table*}

\section{Related Works}

\textbf{Large Vision-Language Models (VLMs).} The great development of Large Language Models (LLMs) has facilitated the community in both academics~\cite{vary,dreamllm,merlin,InstructBLIP,LLM_MiniGPT4} and industries~\cite{ACM_Retrieval,ACM_VQA,VLMevalkit}. Recently, large vision-language models emerged, aiming to extend the reasoning brain of LLM to vision modality~\cite{chatspot}. The majority of VLMs undergo a two-stage training manner, with the pretraining period focusing on aligning the vision and text signals before finetuned with instruction following data in the second stage. LLaVA~\cite{LLM_Llava}, as one of its representatives, has attracted numerous researchers~\cite{DiffusionSFT,ShareGPT4V,LVIS-Instruct4V,VILA} since it provide a valuable opportunity for them to reproduce and built upon. Subsequent works based on LLaVA try to leverage more image-text data pairs of higher quality~\cite{ShareGPT4V} or to construct more abundant instruction following database~\cite{LVIS-Instruct4V}. Despite their achievements, these model are not preference aligned with user intentions neither implicitly nor explicitly, which might limit their further deployment. Therefore, an suitable alignment technique is of great importance.

\textbf{Preference alignment in LLM/VLM.} Training an LLM to align with human or user preference is called RLHF~\cite{InstructGPT}, which usually happens after the instruction following stage. The core concept of RLHF is to reduce un-intented or toxic output produced by LLMs~\cite{SafelyFT,redteaming}. Variants of RLHF include, but not limited to, DPO~\cite{DPO}, PPO~\cite{PPO} and RLAIF~\cite{RLAIF}. Alignment has been fruitfully researched in natural language processing (NLP) domains~\cite{RLAIF,ReST}, but relatively less visited in vision-language fields. There are some latest attempts~\cite{HA-DPO,VLLM_DPO,LLaVa-RLHF} that integrate preference alignment with DPO in vision-language domains. However, they only emphasize on task specific ability (e.g. hallucinations in HA-DPO~\cite{HA-DPO}), and, more importantly, it requires tedious GPT-4 or human interventions to construct the preference data, which may diminish the pipeline efficiency as well. In this paper, we propose an unsupervised data construction pipeline that not only solve the data hungry issue in preference alignment, but also greatly boosts the comprehension ability of current VLMs, as well.

\textbf{Contrastive learning.} Self-supervised learning (SSL) of visual representation are popularized in the past years~\cite{MOCO,SimCLR,MLS,BYOL}. Traditional SSL are mostly contrastive based, where strongly augmented positive views of the same image \emph{will} share similar deep semantic in the hidden space that the model could capture. Nevertheless, we found that a similar augmentation pipeline \emph{does not} holds true in VLMs. Inspired by unsupervised property of SSL, we design an self-supervised pipeline to construct the preference data, and empower the current VLM with improved capability.

\begin{figure*}
	\centering
    \includegraphics[width=0.95\linewidth]{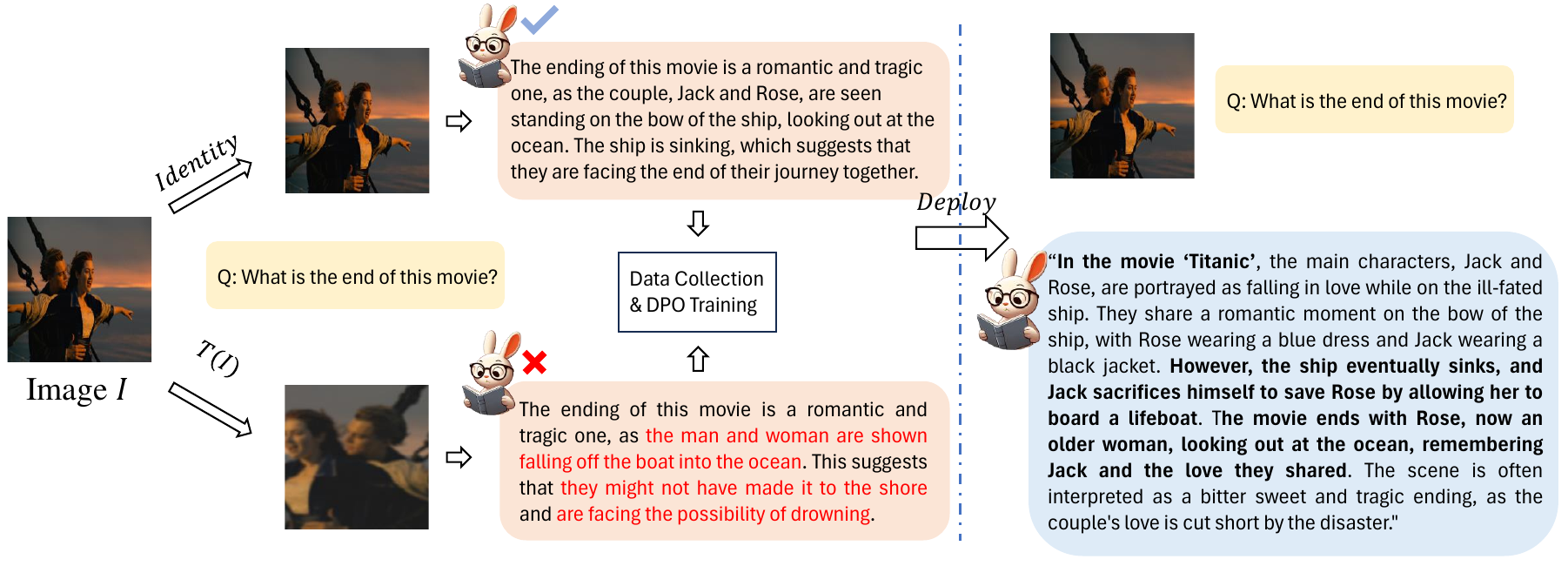}
	\caption{The pipeline of SeVa. For each image $I$ in the selected dataset, we transform it with data augmentation $T$ to obtain the distorted one, while keeping a copy of the original image to form a pair. The shared questions are acted on the paired images to get the chosen and rejected responses, respectively, which undergo a data collection (e.g., filtering) process before the DPO training. In the left part, incorrect words or sentences are red color coded, while in the right part (the improved version of the model), we highlight excellent content with bold phase. Note that in the picture, we show the \emph{same} image both for training and for testing, but actually the data distribution between them are different (\cf Sec.~\ref{sec:exp-settings}). This figure is best viewed in color.}
	\label{fig:method}
\end{figure*}

\section{Method}
We will first show the preliminaries of language modeling and direct preference optimization, then move onto the procedure of the proposed SeVa pipeline. Finally, we discuss its strong relation with visual contrastive learning.

\subsection{Preliminaries} 
\textbf{Language modeling.} We start from the vison-language modeling perspective and take LLaVA~\cite{LLM_Llava} as an example for illustration. Given an image input $I$, it is first passed through a vision encoder (e.g., ViT-L~\cite{CLIP}) to get:
\begin{equation}
    \boldsymbol{h} = g(I) \, ,
\end{equation}
where $g$ denotes a combination of vision encoder and projection layer. $\boldsymbol{h}$ represents a sequence of deep image embeddings. The embeddings before the last transformer layer are considered in LLaVA-1.5~\cite{LLaVa1.5}, which are then concatenated with the question token $q$ and are fed into the auto-regressive LLM $\pi$ that sequentially generates the next target token $y$:
\begin{align}
    &\pi_{\theta} (y|x) = \prod_{i=1}^{L} \pi_{\theta}(y_i|y_{<i},x)\,,
    \label{eq:sequential}
\end{align}
where $x = (\boldsymbol{h}, q)$, and is considered as prompt sent to the LLM parameterized by $\theta$. $y_{<i}$ means the generated token before the current prediction $y_i$, and $L$ is the length of the produced token sequece. We denote the LLM trained with SFT data as $\pi_{\text{SFT}}$ for clarity.

\textbf{Direct preference optimization.} DPO is first proposed in~\cite{DPO}, and can be viewed a new parameterization of the reward model in RLHF~\cite{InstructGPT} that can directly optimize the policy (the LLM's parameter $\theta$). Here we start with RLHF for better illustrations. In RLHF, a Bradley-Terry (BT) reward model is often adopted, which stipulates the human preference distribution as:
\begin{equation}
    p^*(y_{c} \succ y_{r} |x) = \frac{\exp{(r^*(x, y_c))}}{\exp{(r^*(x, y_c))} + \exp{(r^*(x, y_r))}}\,.
    \label{eq:reward}
\end{equation}
The $y_c$ and $y_r$ are the chosen and rejected response conditioned on the LLM's prompt $x$. The reward model $r_{\phi}(\cdot)$ is paramerized by $\phi$ and can be obtained with maximum likelihood using preference database $\mathcal{D}$ annotated by human~\cite{DPO}:
\begin{equation}
    \mathcal{D} = \{x^{(j)}, y_c^{(j)}, y_r^{(j)}\}_{j=1}^N\,.
\end{equation}
Then we maximize the a preference policy as follows:
\begin{equation}
    \max_{\theta'}\mathbb{E}_{x,y}\left\{ r_{\phi}(x, y) - \beta \mathbb{D}_\text{KL}[\pi_{\theta'}(y|x) | \pi_{\text{ref}}(y|x) ] \right\}\,.
    \label{eq:rlhf}
\end{equation}
The reference models $\pi_\text{ref}$ are usually initialized by $\pi_\text{SFT}$, which prevents the learned parameter $\theta'$ from too much deviation. The finally learned policy $\pi_{\theta'}(\cdot)$ are better equipped with human or user intentions.

DPO uses a closed form~\cite{DPO} derived from Eq.~\ref{eq:rlhf} to represent the optimal reward $r^*$ model by the learned optimal preference model $\pi^*$ as (with $K$ the constant factor):
\begin{equation}
r^*(x, y) = \beta \log\frac{\pi^*(y|x)}{\pi_{\text{ref}}(y|x)} + K\,.
\label{eq:form}
\end{equation}
By substituting the $r^*$ in Eq.~\ref{eq:reward} with Eq.~\ref{eq:form}, we get the final optimized loss function of DPO as follows (\cf~\cite{DPO}):
\begin{equation}
    \mathcal{L}_{\text{d}} = -\mathbb{E}_{\mathcal{D}}\left[\log \sigma\left(\beta \log\frac{\pi_{\theta'}(y_c|x)}{\pi_{\text{ref}}(y_c|x)}- \beta\log\frac{\pi_{\theta'}(y_r|x)}{\pi_{\text{ref}}(y_r|x)}\right)\right]
    \label{eq:dpo}
\end{equation}
The trainable parameter Eq.~\ref{eq:dpo} is $\theta'$, which is the same with RLHF. But it do not need any online optimization process like PPO~\cite{PPO}. Instead, DPO samples $(x,y_c, y_r)$ from the preference database $\mathcal{D}$ and optimize the VLM's parameters $\theta'$ with a simple classification loss.

\subsection{Visual preference alignment}\label{sec:method:seva}
This section will introduce our pipeline: self-supervised visual preference alignment. For a language model $\pi_\text{SFT}$ after supervised finetuning stage, we feed it with the original and augmented image input as (for simplicity, we omit the generated token as conditional input shown in Eq.~\ref{eq:sequential}):
\begin{align}
    &y_c^{(j)} = \pi_{\text{SFT}}(g(I^{j}), q^{j})\,, \\
    &y_r^{(j)} = \pi_{\text{SFT}}(g(\mathcal{T}(I^{j})), q^{j})\,.
\end{align}
The $y_c^{(j)}$ and $y_r^{(j)}$ are regarded as a pair of raw chosen and rejected response. For each image $I^j$ and its question $q^j$, we generate only one paired preference data $(y_c^{(j)}, y_r^{(j)})$. Suppose there are total $M$ image-question pairs and its corresponding generated preference pairs, we simply filter those equaled responses to get remained preference data pairs (supposed $N_d$), which constitute the final preference database $\mathcal{D}_\text{SeVa}$ for DPO training (\cf Alg.~\ref{algo:1} and Eq.~\ref{eq:dpo}).
\begin{align}
    \mathcal{D}_{\text{SeVa}} &= \{I^{j}, q^{j}, y_{c}^{(j)}, y_{r}^{(j)}\}_{j=1}^{N_d}\,,\\
    & \text{s.t.}\,\, \forall j,\, y_{r}^{(j)} != y_{r}^{(j)}\,.
    \label{eq:seva-dataset}
\end{align}

We choose 6 different data augmentation strategies to construct our database and conduct DPO training. As shown in Table~\ref{tab:data-aug-improve}, all self-constructed preference data are helpful for multi-modal comprehension, showing the validity of this pipeline. What surprises us is that training samples built from `RandFlip' are quite helpful (e.g., an increase of 3.2\% score on MMVet). We conjecture that this might derive from the natural property of the unsupervised textdata, since flipped images are can make OCR and recognition harder and induce potential negative responses (\cf appendix for more details). Besides, all models have seen a noticeable improvement in adversarial settings of POPE, indicating the models trained by our self-supervised pipeline might be less inferior to potential interruptions.

\subsection{Connection with contrastive learning}\label{sec:method:ssl}
In contrastive learning, augmented views of the same image will share similar semantic in its deep embeddings, which optimize the following InfoNCE loss:
\begin{equation}
\mathcal{L}_\text{in} = - \log \frac{\exp{ (q\cdot k_+/\tau)}}
{\exp{(q \cdot k_+/\tau)} + \sum_{i}^{n} \exp{(q \cdot k_-^i / \tau)}} \,,
\label{eq:infonce}
\end{equation}
in which $q$, $k_+$ are the positive embeddings from augmented views of the same image, while $k_{-}$ could be negatives embeddings from other images~\cite{SimCLR} or memory bank~\cite{MOCO}. If we consider only one negative, and denote $f(q,k) =(q\cdot k)/\tau$ as the scoring function, we can reformulate Eq.~\ref{eq:infonce} as:
\begin{equation}
    \mathcal{L}'_\text{in} = - \log \frac{\exp{(f(q,k_+})) }
{\exp{(f(q,k_+))} +  \exp{(f(q,k_-))}} \,.
\label{eq:infonce2}
\end{equation}
It is not hard to find that the optimzed loss function derived from Eq.~\ref{eq:reward} are quite similar with Eq.~\ref{eq:infonce2}. The main difference lies in the \emph{definition of negative samples}. Unlike contrastive learning, the predicted tokens from the augmented views of the same image will be counted as negatives, but not positives. If we imitate traditional InfoNCE loss in Eq.~\ref{eq:infonce}, we could inject multiple negative reward terms in Eq.~\ref{eq:reward} (by augmenting multiple views of the same image to produce multiple responses), and thus derive a more general form of DPO. Please refer to appendix for more details, and we will leave this as future work.

\section{Experiment}
In this section, we conduct experiments to verify the effectiveness of the proposed SeVa pipeline. Firstly, we provide the experimental settings (e.g., the data construction process). Then we show the results on multi-modal comprehension benchmarks. Finally, we provide fruitful ablations and visualizations to analyze SeVa with more details. For simplicity, \emph{we name LLaVA-1.5-7B/13B trained with SeVa pipeline as SeVa-7B, SeVa-13B respectively.}

\subsection{Settings}\label{sec:exp-settings}
\textbf{Data construction.} The source data we obtained are from LLaVA665k SFT dataset~\cite{LLaVa1.5}, and we choose image-question pairs from TextVQA~\cite{TextVQA} and OCRVQA~\cite{OCRVQA} (denoted as `text+ocr' for simplicity) in LLaVA665k to generate the DPO preference data. \emph{This setting is kept by default throughout all the experiment unless otherwise noted}. For each data instance in `text+ocr' of LLaVA665k, we randomly choose 2 questions in its multi-turn dialogue, and pair each question with the image. We treat each image-question pair as \emph{one data instance}, which leads to nearly doubled instance as the original `text+ocr'. Then we randomly choose 8k such image-question pairs in `text' and `ocr' each, to form the original data source of 16k. Finally, these 16k image-question data instances will go through SeVa pipeline to produce DPO preference data with filtering (\cf Alg.~\ref{algo:1}), which are cut down by half to about 8k. For clarity in our later ablations, we denote training data num as those unfiltered preference pairs instead of the filtered ones (e.g., 16k but not 8k), since the raw unfiltered data is fixed and immune to filtering process (\cf appendix). 

\textbf{Data augmentation choice.} As verified in Table~\ref{tab:data-aug-improve}, all the selected data augmentations are all helpful for multi-modal comprehension. In our later experiment, we choose \emph{diffusion noise} as the default augmentation in SeVa pipeline, since it can easily control the intensity of distorted level, which is helpful for ablations (\cf Fig.~\ref{fig:hard-neg}). The noise steps for training SeVa-7B and SeVa-13B are set as 800 and 500, respectively. As MOCO augmentation strategy is also highly effective as shown in Table~\ref{tab:data-aug-improve}, we are glad to verify more data augmentation strategies in our future work.

\begin{table*}
	\small
	\centering
	\begin{tabular}{llllllllllll}
		\toprule[1pt]
		 Method & Language model  & MMVet & LLaVA$^\text{W}$  & MMB & MMB$^\text{CN}$ & POPE & SEED$^{I}$ & SHR $(\downarrow)$ & SQA & GQA \\
		\midrule[1pt]   
        BLIP-2 & FLAN-T5 & 22.4 & 38.1 & -- & -- & 85.3 & 46.4 & -- & 61.0 & 41.0\\
        InstructBLIP & Vicuna-7B & 26.2 & 60.9  & 36.0 & 23.7  & -- & 53.4 & - & 60.5 \\
        InstructBLIP & Vicuna-13B & 25.6 & 58.2  & -- & --& 78.9 & -- & 51.2 & 63.1 & 49.5\\
        Shikra & Vicuna-13B & -- & -- & 58.8 & -- & -- & -- & -- & -- & --   \\
        IDEFICS-9B & LLaMA-7B & -- & --  & 48.2 & 25.2 & -- & -- &  -- & -- & 38.4 \\
        IDEFICS-80B & LLaMA-65B & -- & -- & 54.5 & 38.1 & -- & -- &  -- & -- & 45.2 \\
        Qwen-VL & Quen-7B & -- & -- & 38.2  & 7.4 &  -- & 56.3 & -- & 67.1 & 59.3 \\
        Qwen-VL-chat & Quen-7B & -- & --  & 60.6 & 56.7 & -- & 58.2 & -- & 68.2 & 57.5 \\
        LLaVA & Vicuna-7B & 26.7 & 63.0 & 34.1 & 14.1 & -- & 25.5 & -- & 38.5  & --\\
        \midrule
        LLaVA-1.5 & Vicuna-7B & 30.5 & 63.4  & 64.3 & 58.3 & 85.9 & 65.7  & 36.7 & 66.8 & \textbf{62.0} \\
        SeVa-7B & Vicuna-7B & \textbf{37.2} & \textbf{72.2}  & \textbf{65.6} &  \textbf{59.2}& \textbf{86.7} & \textbf{65.8}  & \textbf{34.9} & \textbf{67.5} & \underline{60.7} \\
        \midrule
        LLaVA-1.5 & Vicuna-13B & 35.4 & 70.7& 67.7 & 63.6 & 85.9 & 68.2 & 37.2 & \textbf{71.6} & 63.3 \\
        SeVa-13B & Vicuna-13B & \textbf{41.0} & \textbf{80.1} & \textbf{68.7} & \textbf{64.8} & \textbf{87.4} & \textbf{68.6} & \textbf{36.6} & \underline{71.2} & \textbf{63.4}\\
        \bottomrule[1pt]
	\end{tabular}
\caption{Comparison with state-of-the-art methods on 9 benchmark datasets.  SeVa consistently improves LLaVA-1.5-7B/13B on 8 out of 9 benchmarks, namely, MMVet~\cite{MMVet}, LLaVA-bench-in-the-wild~\cite{LLM_Llava}, MMBench~\cite{MMBench}, MMBench-Chinese~\cite{MMBench}, POPE~\cite{POPE}, SEED-Image~\cite{Seed-Bench}, SHR~\cite{HA-DPO}, SQA~\cite{GQA}, GQA~\cite{GQA}. Among them, SHR are newly proposed~\cite{HA-DPO} to evaluate the hallucination ratio of VLMs.}
\label{tab:compare-with-sota}
\end{table*}

\begin{table*}
	\small
	\centering
	\begin{tabular}{lccccccc}
		\toprule[1pt]
		Method &  Unsupervised& data source & data scale &  MMVet & MMB & POPE & Training cost \\
        \midrule
        \textcolor{lightgray}{LLaVA-1.5} & \textcolor{lightgray}{---} & \textcolor{lightgray}{---} & \textcolor{lightgray}{---} & \textcolor{lightgray}{30.5} &  
 \textcolor{lightgray}{64.3} & \textcolor{lightgray}{85.9} & \textcolor{lightgray}{---}\\
        cont. SFT (2\%)  & \ding{55} & LLaVA665k & 15k & 31.9 & 63.7 & 86.0 & 0.1h \\
        cont. SFT (10\%)  & \ding{55} & LLaVA665k & 66k & 32.8 & 64.9 & 86.0 & 1.2h \\
        cont. SFT (20\%)& \ding{55} & LLaVA665k& 132k & 33.9 & 64.2 & 86.1 & 2.4h\\
        cont. SFT  & \ding{55} & text+ocr  & 102k & 32.5 & 65.2 &  86.7 & 1.2h \\
        \midrule
        SeVa (\emph{ours})& \ding{51}  & text+ocr & 8k$^*$ & 34.8 & 65.3 & 86.2 & 0.3h\\
        SeVa (\emph{ours})& \ding{51}  & text+ocr & 16k$^*$ & \textbf{37.2} & \textbf{65.6} & \textbf{86.7} & 0.7h\\
        \bottomrule[1pt]
	\end{tabular}
\caption{Comparison between SeVa with continual (`cont.') SFT in a 7B setting. We randomly select 15k (2\%), 66k (10\%) and 132k (20\%) out of LLaVA665k as SFT data. `Unsupervised' means no answering labels are required. We also sample \emph{all} the TextVQA and OCRVQA data in LLaVA665k for comparison (since SeVa utilize its unsupervised data version). $^*$ means the unfiltered preference instances (the actually kept data for DPO training are mostly by half, \cf Sec.~\ref{sec:exp-settings}). Training cost are all evaluated in a same experiment settings.}
\label{tab:ablate-sft}
\end{table*}

\textbf{Training details.} Following previous works~\cite{ShareGPT4V,LVIS-Instruct4V,LLaMA-VID,DiffusionSFT}, we choose LLaVA-1.5-7B/13B as our base models in all our experiments. The LLaVA's weights are pretrained and SFT tuned before our DPO training, which is the by default pipeline as in many previous literatures~\cite{HA-DPO,ReST,DPO,PPO}. We adopt deepspeed ZeRO stage-3~\cite{LLaVa1.5} during DPO training and use Vicuna-7B/13B, CLIP-VIT-L-336px as our LLM and vision encoder, respectively. The total epoch, batch size, learning rate and weight decay are set as 1, 128, 2e-6 and 0, respectively, following previous work~\cite{HA-DPO}.  The hyper-parameters of lora $r$ are set as 1024, and the scale parameter $\beta$ in DPO is fixed as 0.1. DPO optimization are conducted with 8-A800 GPU, with LLaVA-1.5-7B/13B tuned for about 0.7/1.3 hours.

\subsection{Compare with state-of-the-art}
Firstly, we compare the proposed SeVa pipeline with state-of-the art VLMs. A total of 9 benchmarks are included, with multi-modal benchmark as: MMVet~\cite{MMVet}, LLaVA-Bench-in-the-wild~\cite{LLaVa1.5} (LLava$^\text{W}$), MMBench~\cite{MMBench} (MMB), MMBench-Chinese (MMB$^\text{CN}$), POPE~\cite{POPE}, SEED-Bench-Image~\cite{Seed-Bench} (SEED$^I$) and SHR~\cite{HA-DPO}, and traditional Question-Answer benchmarks ScienceQA~\cite{SQA} (SQA) and GQA~\cite{GQA}. Among them, MMVet, LLaVA$^\text{W}$ are two GPT-4 evaluated benchmarks. MMB and MMB$^\text{CN}$ are multiple choice question answer dataset that requires minimal GPT-4 involvement (e.g., answer choice aligning~\cite{MMBench}). POPE and SHR are two hallucination benchmarks, and SQA \& GQA are two traditional QA benchmarks.

As shown in Table~\ref{tab:compare-with-sota}, SeVa consistently improve the LLaVA-1.5 7B/13B models on 8 out of 9 benchmark datasets. Specifically, it improve LLaVA-v1.5-7B/13B models by 6.7\% and 5.6\% on the most complex multi-modal dataset MMVet, showing the great power the proposed pipeline. The same can be observed in LLaVA$^\text{W}$, where a 13B SeVa model achieves a 80\% relatively score to GPT-4, boosting LLaVA-1.5-13B by a large margin of 9.4\%. The great boost on GPT-4 involved datasets suggested better alignment with user-intentions, as demonstrated in Fig.~\ref{fig:win} and Fig.~\ref{fig:question-answer}. We also observe a consistent improve on benchmarks POPE and SHR, which indicates the SeVa pipeline is also helpful for relieving object hallucinations, and thus we \emph{may not} need specialized hallucination methods~\cite{VCD,HA-DPO} to handle them. Please note that the improvement on hallucinations are \emph{not marginal}, since previous work that constructed huge amount of SFT data can lead to \emph{decreased performance} on POPE (e.g., LVIS-Instruct4V~\cite{LVIS-Instruct4V}). There is also a higher multiple choices accuracy in MMB, MMB$^\text{CN}$ and SEED$^I$, showing the broad adaptability of SeVa. In Table~\ref{tab:compare-with-sota}, a little bit drop was found on traditional benchmarks SQA/GQA. We conjecture that this phenomenon might arise from a trade-off between stronger instruction following ability in traditional QA and better comprehension ability in modern VLMs benchmarks, which is also pointed out in~\cite{ReST} that stronger instruction following do not guarantee a more helpful and better VLMs. In fact, how to better trade-off between benchmarks remains a discussing issue~\cite{ShareGPT4V} and is out-of-scope of this paper.

\begin{table}
	\small
	\centering
	\begin{tabular}{cllccccc}
		\toprule[1pt]
		  \multirow{2}{*}{Unspervised data} &\multicolumn{4}{c}{Data Scale} \\
            & 2k & 4k & 8k & 16k \\
        \midrule

       TextVQA~\cite{TextVQA}  & 31.8 & 32.1 & 34.8 & \textbf{35.8} \\
        OCRVQA~\cite{OCRVQA} & 32.1& 32.3 & 32.8 & \textbf{34.5}  \\
        GQA~\cite{GQA}  & 31.4 & 31.8 & 34.1 & \textbf{35.9} \\
        COCO~\cite{MOCO} & 31.7 & 31.2 & 34.2 & \textbf{34.4} \\
        Visual Genome~\cite{LLaVa1.5} & 31.4 & 32.0 & 33.4 & \textbf{34.1} \\

        \bottomrule[1pt]
	\end{tabular}
\caption{MMVet evaluation score with different data (image-question) pairs and scale to conduct the SeVa-7B pipeline. The highest score in each dataset are highlighted with bold symbol. In SeVa pipeline, we choose TextVQA+OCRVQA by default. All the data listed are sourced from LLaVA665k~\cite{LLaVa1.5}.}
\label{tab:ablate-dataset}
\end{table}

\begin{figure}
	\centering
    \includegraphics[width=0.85\linewidth]{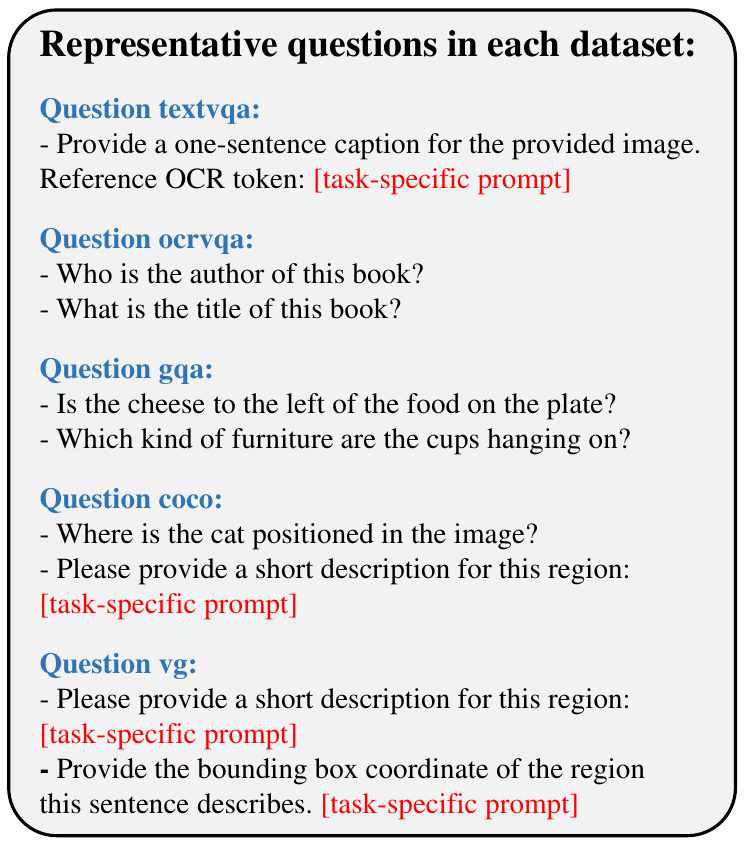}
	\caption{Illustration of representative questions in five different database from LLaVA665k~\cite{LLaVa1.5}. In our main experiment, we adopt a combination of `textvqa' and `ocrvqa'. The results of applying the other 3 question types in SeVa can be found in Table~\ref{tab:ablate-dataset}.}
	\label{fig:dataset_question}
\end{figure}

\begin{table}
	\small
	\centering
	\begin{tabular}{lllllll}
		\toprule[1pt]
		\multirow{2}{*}{temp ($t$)} & \multicolumn{2}{c}{Q-Consistency ($\uparrow$)} &\multicolumn{2}{c}{A-Consistency ($\uparrow$)} \\
        &LLaVA &  SeVa & LLaVA & SeVa \\
        \midrule
        0.2 & 7.30 & \textbf{7.75} & 6.45 & \textbf{6.95}\\
        0.4 & 7.11 & \textbf{7.98} & 6.28 & \textbf{7.32} \\
        0.5 & 7.39 & \textbf{7.93} & 6.52 & \textbf{7.29}\\
        0.7 & 7.81& \textbf{8.18} & 7.10 & \textbf{7.64}\\
        0.9 &8.43 & \textbf{8.53} & 7.87 & \textbf{7.98}\\
50 & 7.75 & \textbf{8.47} & 6.94 & \textbf{8.03}\\
        \bottomrule[1pt]
	\end{tabular}
\caption{GPT-4 evaluated consistency score from 1-10. Q-consistency measures how the model's answers are aligned with the asked question, while A-consistency measures how answers ($t>0$) are aligned with its generated tokens w/o sampling ($t=0$).}
\label{tab:ablate-consistency}
\end{table}

\subsection{Ablations}
Now we move onto ablating the factors in the proposed pipeline, to help readers better understand the success behind SeVa and its generalized ability. We choose SeVa-7B as the default settings unless otherwise noted.

\textbf{Compare with SFT} We start from comparing with continual SFT, as this is the most straight forward method to verify its effectiveness with supervised training (note SeVa \emph{do not} need target answer labels). We conduct three different types of SFT: a random selection of 2\%/10\%/20\% in LLaVA665k and a collection of TextVQA and OCRVQA in 665k (denoted as `text+ocr'). We evaluate its downstream performance on MMVet, MMB and POPE, which can be found in Table~\ref{tab:ablate-sft}. For a fair comparison, we strictly follow LLaVA-1.5's SFT settings in experiments. It can be observed that all SFT data are helpful for multi-modal comprehension. Specifically, text related data are more helpful for dealing with model hallucinations (e.g., on POPE). In comparison, our SeVa achieves the overall best results on all benchmarks with both less training time cost and data usage. More importantly, SeVa enjoys its unsupervised nature, making it more generalizable in reality.

\begin{figure}
	\centering
    \begin{subfigure}{0.49\linewidth}
		\includegraphics[width=0.99\linewidth]{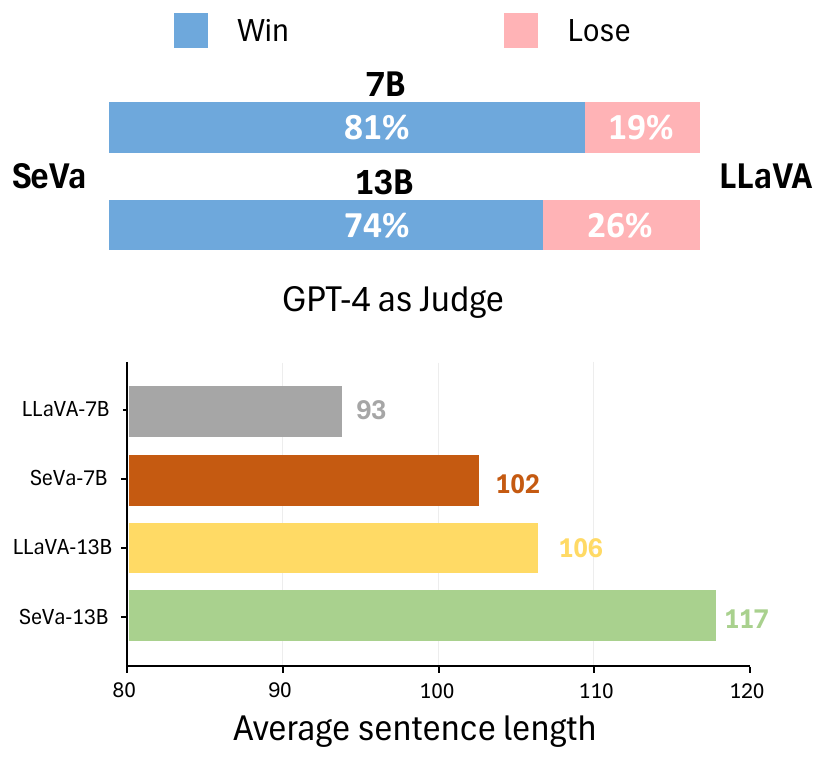}
		\caption{results on LLaVA$^\text{W}$}
		\label{fig:win-llavabench}
	\end{subfigure}
	\begin{subfigure}{0.49\linewidth}
		\includegraphics[width=0.99\linewidth]{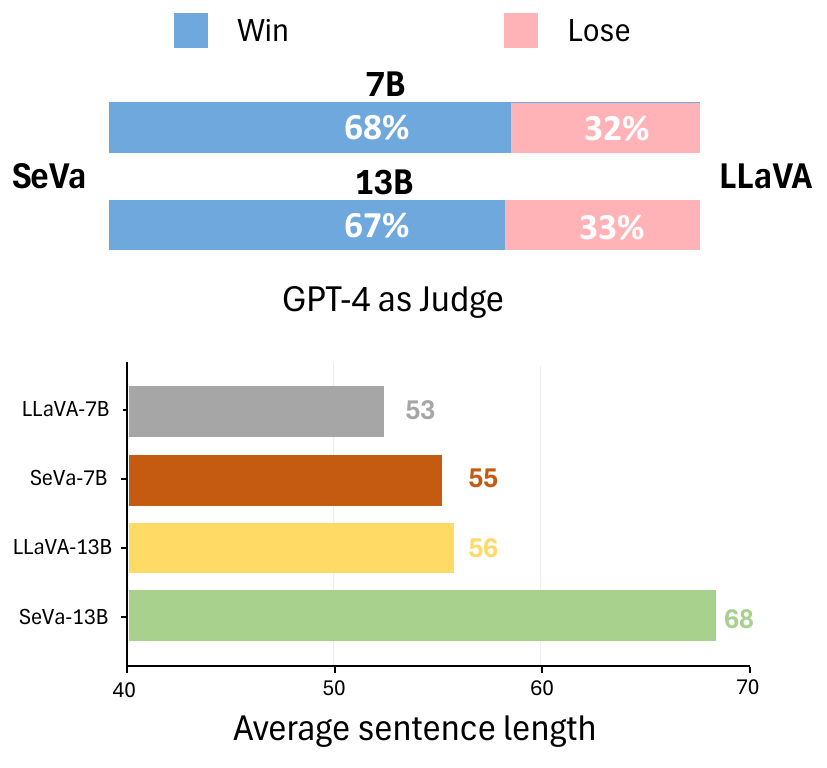}
    	\caption{results on MMVet}
		\label{fig:win-mmvet}
    \end{subfigure} 
	\caption{The pair-wise competition and output sentence length (by token) in LLaVA$^\text{W}$ and MMVet, respectively. We compare between SeVa and LLaVA-1.5 models in 7B and 13B settings.}
	\label{fig:win}
\end{figure}

\begin{figure*}
	\centering
    \includegraphics[width=0.99\linewidth]{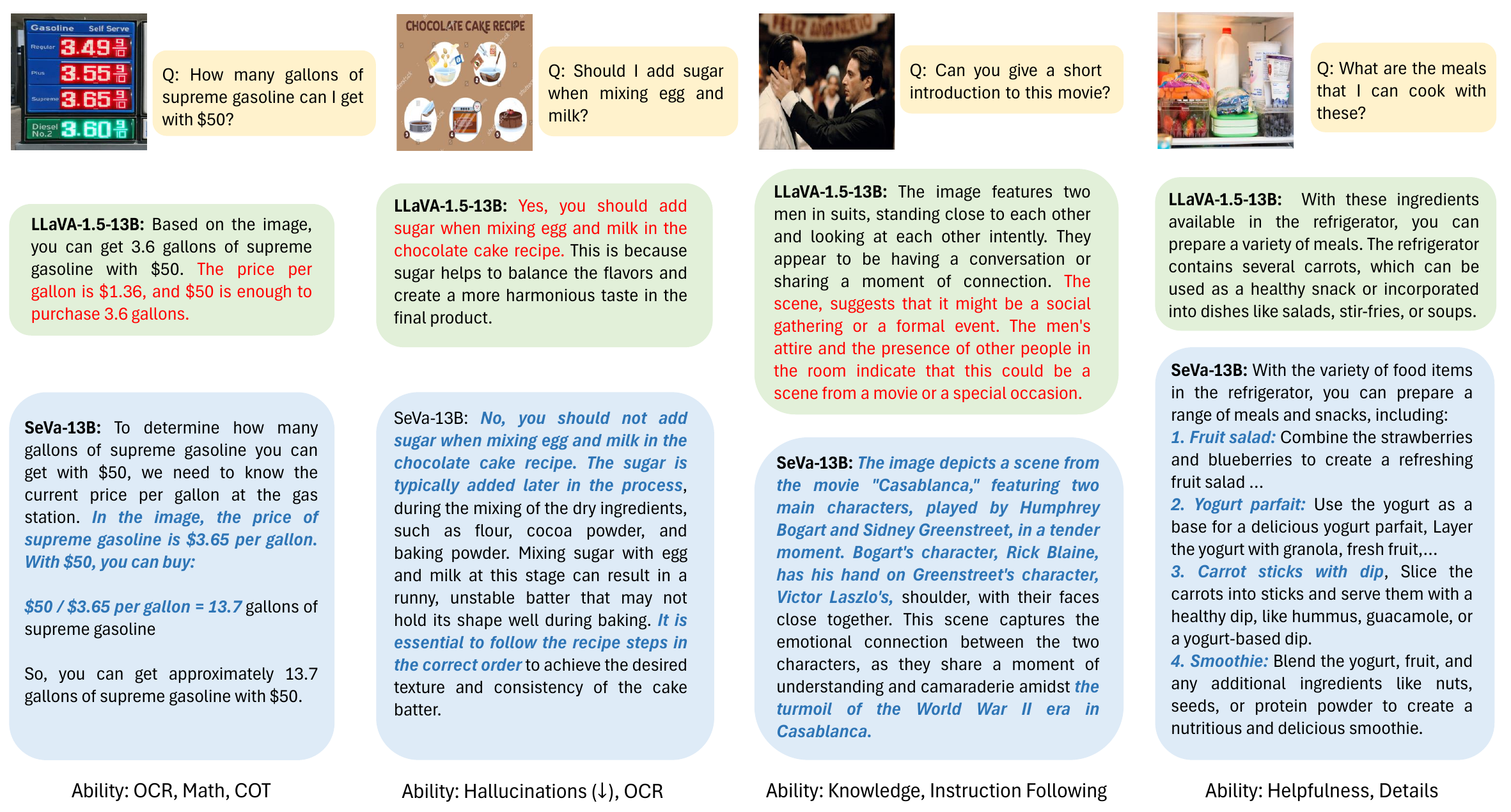}
	\caption{Four examples to illustrate the improved ability of our SeVa, including chain of thought (COT), stronger OCR, less hallucinations, world knowledge and more detailed and clearer responses. Since we built-on LLaVA-1.5, we directly make comparison between them to show the effectiveness of our proposed method. For clarity, we highlight the incorrect response in LLaVA-1.5 with red color, and emphasize ours with blue italic character. This figure is best viewed in color.}
	\label{fig:question-answer}
\end{figure*}

\begin{figure}
	\centering
    \begin{subfigure}{0.49\linewidth}
		\includegraphics[width=0.99\linewidth]{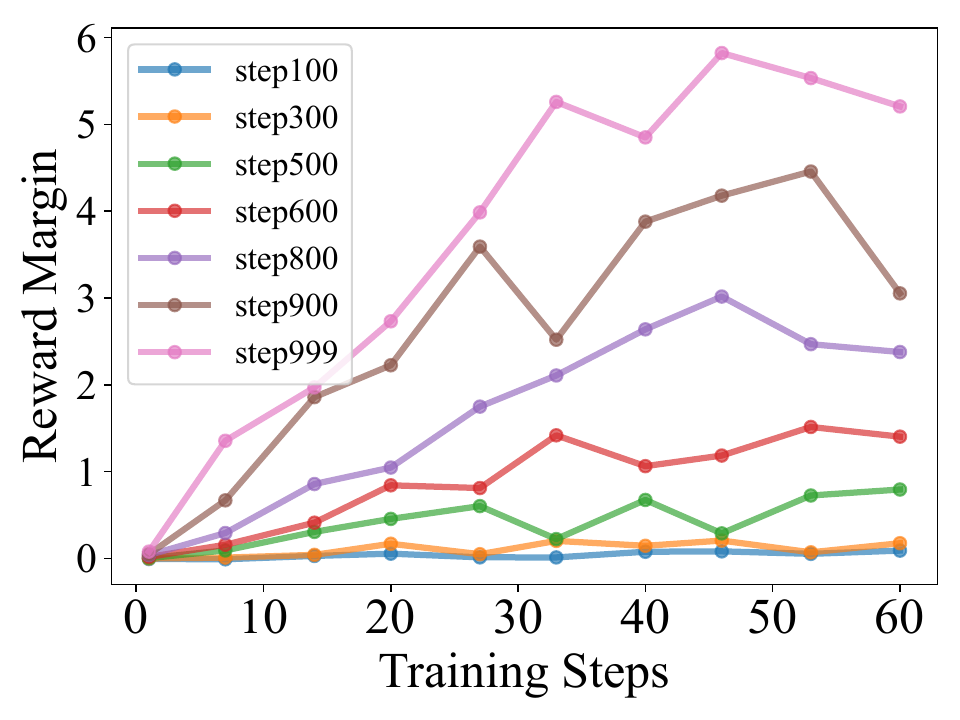}
		\caption{reward magin of DPO}
		\label{fig:hard-neg:margin}
	\end{subfigure}
	\begin{subfigure}{0.49\linewidth}
		\includegraphics[width=0.99\linewidth]{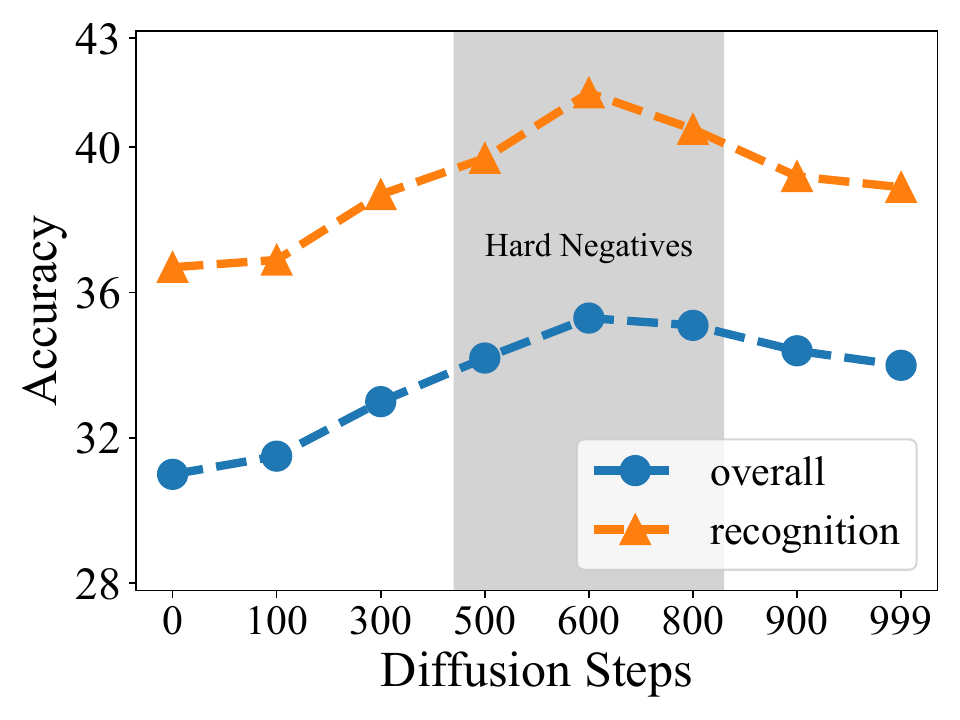}
    	\caption{overall score on MMVet}
		\label{fig:hard-neg:mmvet}
    \end{subfigure} 
	\caption{Exploration of hard-negatives. We plot the DPO reward margin between chosen and rejected samples with regard to different diffusion noise steps in data generation process (\cf~\ref{fig:hard-neg:margin}). We conduct DPO with preference data from diffusion noise augmentations, and show their evaluation scores on MMVet (\cf~\ref{fig:hard-neg:mmvet}).}
	\label{fig:hard-neg}
\end{figure}

\textbf{Generalize to other datasets.} Since we adopt the questions in TextVQA and OCRVQA as a default fashion, we now ablate on more dataset selection on top of these two dataset. For more clarity, we choose other part of LLaVA665k including GQA part, COCO part and Visual Genome part to form the image-question pairs. We randomly select 2k, 4k, 8k and 16k raw image-question pairs and conduct SeVa pipeline to collect data and apply DPO training with 7B models. The results are then evaluated on MMVet. As shown in Table~\ref{tab:ablate-dataset}, all the choosen dataset lead to consistent improvement over LLaVA-1.5-7B (30.5\% score), showing the generalized ability of SeVa to different image and question domains. Besides, involving more unsupervised data leads to better performance boost, which indicates the great power of SeVa towards further scaling.

\textbf{Why SeVa works?} Attentive readers might have a natural question: \emph{why rejecting inferior answers lead to better model output than the original?} Here, we conduct a detailed experiment to show that, SeVa can be viewed as a special form of model calibration and rejecting negatives \emph{does help}. Specifically, we collect image-question pairs from LLaVA$^\text{W}$ to prompt LLaVA and SeVa with higher generation temperatures $t$ for sampling. Note that $t$=0 means no sampling in token generation, which is the default evaluation settings and is most stable. Then we utilize GPT-4 (the detailed prompt can be found in appendix) to evaluate the its output consistency with scoring from 1 to 10, namely Q-consistency (how the model's answer are aligned with the input questions) and A-consistency (how the model's answer are aligned with its generated tokens w/o sampling). As shown in Table~\ref{tab:ablate-consistency}, the consistency score of both models fluctuates as $t$ grows. However, SeVa has seen an consistent better score in all different $t$s, especially when temperature becomes higher. Since VLMs are easily affected by randomness in generating tokens~\cite{VLLM_DPO}, SeVa has potentially made a calibrating effect by reducing the sampled probability of those \emph{bad} tokens, thus being more tolerable to interruption (e.g., temperatures). As a result, the model trained with SeVa can produce more robust and correct answers.

Besides, we have also observed a longer response in LLaVA$^\text{W}$ and MMVet (\cf Fig.~\ref{fig:win}). This indicates that models trainied with SeVa pipeline could potentially produce more detailed and meaningful responses that greatly benefits multi-modal comprehension. 

\textbf{Hard negatives matter.} As we previously observed the phenomenon of hard negatives in Table~\ref{tab:data-aug-improve}, we now design a controllable experiment to study how strength of data augmentation affect the final training performance. Specifically, diffusion gaussian noise with different steps are chosen to generate preference data for DPO training, which are then evaluated on MMVet benchmark. As shown in Fig.~\ref{fig:hard-neg:margin}, larger noise steps will increase the reward margin (\cf DPO~\cite{DPO} for more details) during training, indicating the increased preference divergence. In the meanwhile, a best trade-off exists in the noise steps (possibly between 500 and 800, as shown in Fig.~\ref{fig:hard-neg:mmvet}). We thus attribute the preference samples produced within this regime as \emph{hard negatives} that are most valuable for DPO training.

\begin{table}
	\small
	\centering
	\begin{tabular}{lllll}
		\toprule[1pt]
		 LoRA $r$ & LoRA $\alpha$ &  MMVet & MMB  & POPE \\
        \midrule
        \textcolor{lightgray}{---} & \textcolor{lightgray}{---} & \textcolor{lightgray}{30.7} &  
 \textcolor{lightgray}{64.3}& \textcolor{lightgray}{85.9}\\
        64 & 128 &  32.8 & 64.7  & 86.5 \\
        128 & 256 & 33.9 & 65.0  & 86.4 \\
        256 & 512 & 34.0 & 65.1  & 86.6\\
        512 & 1024 & 35.5 & 65.5 & \textbf{86.8}\\
        1024 & 2048 & \textbf{37.2} & \textbf{65.6} & 86.7\\
        2048 & 4096 & 33.5 & 65.0  & 84.5\\
        \bottomrule[1pt]
	\end{tabular}
\caption{The effect of hyper-parameter in LoRA~\cite{LoRA} during our DPO training. We range the rank of $r$ from 64 to 2048, while keeping the weight ratio fixed as 2 ($\alpha$ is always twice as $r$).}
\label{tab:ablate-lora}
\end{table}

\textbf{Effect of Lora.} For completeness, the hyper-parameters of LoRA's low rank are ablated on three multi-modal comprehension benchmarks, which are shown in Table~\ref{tab:ablate-lora}. In PETL~\cite{DTL}, this parameter can be quite important, as it decides how much new knowledge the model can be absorbed during finetuning. As shown in the table, a relatively higher $r$ is helpful to downstream tasks, showing the increased acquired knowledge of the VLM. However, too larger $r$ will decrease the model performance, which can be attributed to the catastrophic forgetting in training a LLM (a similar point also pointed out in ShareGPT4V~\cite{ShareGPT4V}). In conclusion, we set $r$ as 1024 to seek a best trade-off, and always keep the delta importance as 2 (the $\alpha$ is always twice as $r$).

\textbf{Aligning with user-intentions.} Finally, we visualize the improved VLM with SeVa. We selected image-question pairs from MMVet and LLaVA$^\text{W}$, which are sent to the original VLM (LLaVA-1.5-7B) and the improved model (SeVa-7B), respectively. As shown in Fig.~\ref{fig:question-answer}, Our SeVa demonstrates superiority over LLaVA in various aspects: stronger OCR ability, where SeVa could recognize the exact number of the gas price; less hallucinations can be observed in the second picture, where SeVa accurately comprehends the process of `chocolate cake recipe' and produces the correct answers. It is also surprising that SeVa could potentially recovers more world knowledge after DPO training, as it gives detailed and accurately response to the introduction of a movie (the third picture). Finally, we found that through our DPO training, the models provide more detailed and helpful answers, as suggested by the last picture of a meal ordering. Following Vicuna~\cite{vicuna2023}, we also conduct a competition game between SeVa and LLaVA-1.5 under 7B and 13B settings on LLaVA$^\text{W}$ and MMVet, respectively. We introduce GPT-4 as judge to evaluate the score of each model's responses (similar to the evaluation process in these two benchmarks) and calculate the statistics of `win' and `lose'. \emph{This could serve as a results that indicates how the models are aligned with user-intentions}~\cite{LLM-as-judge}. Note we didn't include `tie'~\cite{vicuna2023}, since we observed a majority of meaningless `tie', where both SeVa and LLaVA achieve a a score of \emph{zero}. As shown in Fig.~\ref{fig:win}, SeVa has occupied most of the winning rounds in comparison with LLaVA, across both model sizes and benchmark datasets. Together with the visualizations in Fig.~\ref{fig:question-answer}, we believe that our SeVa could probably serve as a strong pipeline to improve current VLM's' chat ability, making them more suitable towards practical usage.

\section{Conclusions and future work}
In this paper, we introduce SeVa: Self-supervised visual preference alignment, for multi-modal comprehension. It first went through a data collection and filtering process with response from the original and distorted image, respectively. Then, the standard DPO training are applied to improve the model's capability. Experiments in various benchmarks clear verify and efficiency of the approach. We also conduct ablations and quantitative visualizations to reveal the latent mechanism of SeVa, which demonstrate the great merit of aligning large models in vision-language domains.

In the future, we might consider generalize SeVa to various domains. We will study the effect of data scale: how the model will benefit from DPO training if more unlabeled data are involved, On top of that, we would investigate the potential to generalize our SeVa pipeline RLHF~\cite{InstructGPT} domain. (e.g., whether the generated preference data could be helpful in building a reward model).

\section{Acknowledgement}
We thank Yin-Yin He (M.Sc. degree in Nanjing University, now in ByteDance) for his helpful connections. We also appreciate the discussions with Jie Shao (Ph.D student in Nanjing University) about LLMs.

\appendix

\begin{figure*}
	\centering
    \includegraphics[width=0.88\linewidth]{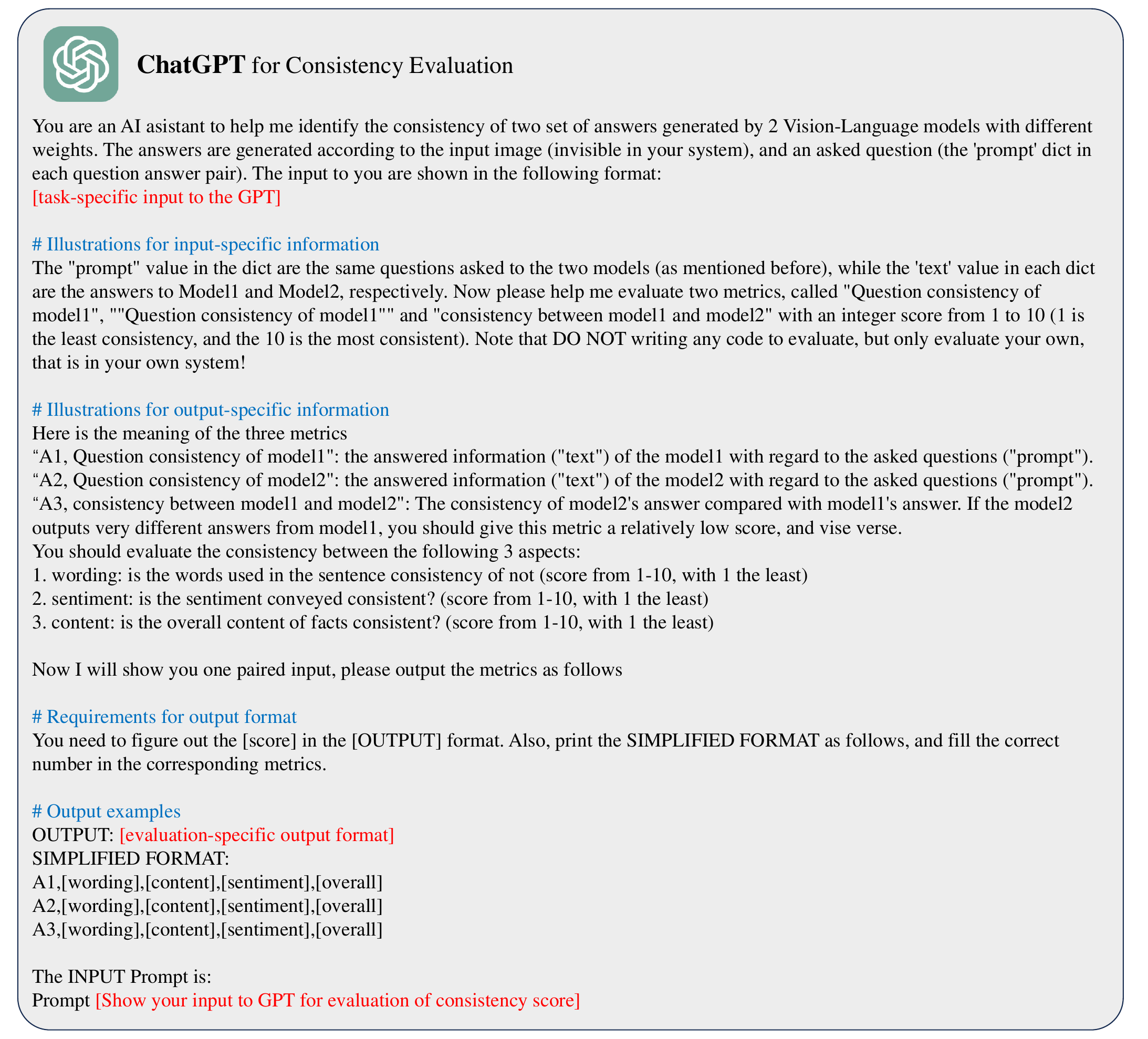}
	\caption{Prompt template to evaluate the consistency score. The numerical results are show in Table~\ref{tab:ablate-consistency}.}
	\label{fig:question-answer-appendix}
\end{figure*}

\section{Relations with contrastive learning}
As discussed before in Sec.~\ref{sec:method:ssl}, the optimized function in SeVa has strong relations with visual contrastive learning. In this section, we will derive a more general form of DPO loss that can be easily adapted to our SeVa pipeline.
We first rewrite the InfoNCE loss here for clarity:
\begin{equation}
\mathcal{L}_\text{in} = - \log \frac{\exp{ (q\cdot k_+/\tau)}}
{\exp{(q \cdot k_+/\tau)} + \sum_{i}^{n} \exp{(q \cdot k_-^i / \tau)}} \,,
\label{eq:infonce-appendix}
\end{equation}
Inspired by Eq.~\ref{eq:infonce-appendix} that multiple negative terms are involved, we can reformulate the preference distribution in Eq.~\ref{eq:reward} as:
\begin{align}
    p^*_{\text{multi}}& = p^*(y_{c} \succ \boldsymbol{Y_{r}} |x) \\
    &= \frac{\exp{(r^*(x, y_c))}}{\exp{(r^*(x, y_c))} + \sum_{y_r \in \boldsymbol{Y_{r}}} \exp{(r^*(x, y_r))}}\,,
    \label{eq:reward-appendix}
\end{align}
where one positive preference data is paired with multiple rejected samples that are represented by a union $\boldsymbol{Y_r}$. In this case, the preference database changes to:
\begin{equation}
    \mathcal{D}^\text{multi} = \{x^{(j)}, y_c^{(j)}, \boldsymbol{Y_r}^{(j)}\}_{j=1}^{N_d}\,.
\label{eq:dpo-data-appendix}
\end{equation}
Note that in SeVa pipeline, this new preference database in Eq.~\ref{eq:dpo-data-appendix} can be easily constructed (e.g., we can do data augmentation sampling across range of available ones to obtain more negatives for $\boldsymbol{Y_c}$). Now, by considering the closed form in of the reward function in Eq.~\ref{eq:form} and utilizing samples in Eq.~\ref{eq:dpo-data-appendix} (to do maximum likelihood estimation), we derive a more general form of DPO as:
\begin{align}
\mathcal{L}_{\text{dpo}}^\text{multi} &= -\mathbb{E}_{\mathcal{D}^\text{multi}}\left[\log \sigma\left(\beta \log\frac{\pi_{\theta'}(y_c|x)}{\pi_{\text{ref}}(y_c|x)}\right.\right. \nonumber \\
&\quad \left.\left.- \sum_{y_r\in\boldsymbol{Y_r}}\beta\log\frac{\pi_{\theta'}(y_r|x)}{\pi_{\text{ref}}(y_r|x)}\right)\right]\,,
\label{eq:dpo-appendix}
\end{align}
and optimize it using pseudo-constructed database by our SeVa pipeline. At present, this generalized form is out of the scope of this paper. We will leave this as future work to explore the potentials of our SeVa.

\begin{figure*}
	\centering
    \begin{subfigure}{0.46\linewidth}
		\includegraphics[width=0.99\linewidth]{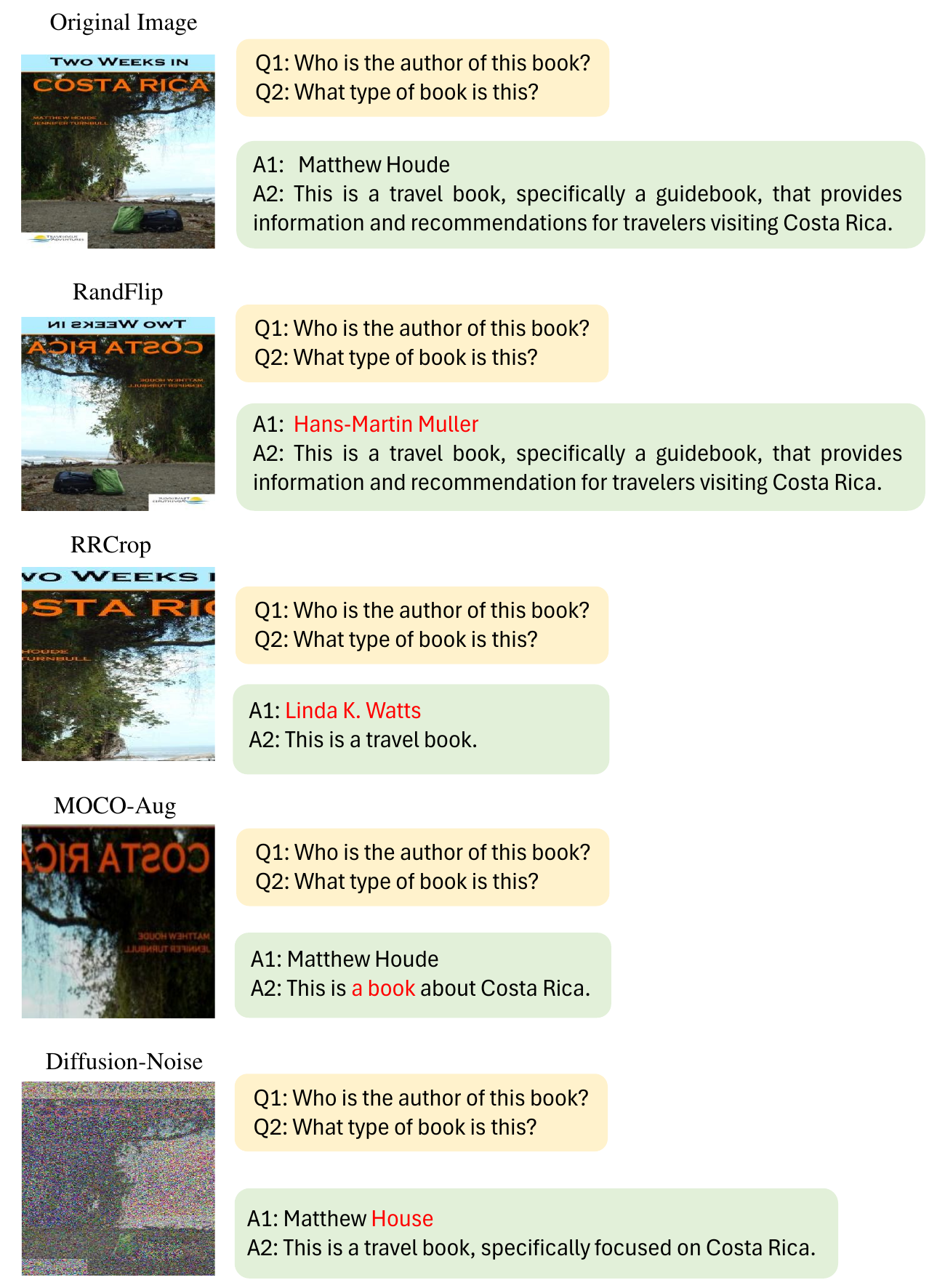}
		\caption{data augmentations of OCRVQA~\cite{OCRVQA} images}
		\label{fig:}
	\end{subfigure}
    \hspace{10pt}
	\begin{subfigure}{0.46\linewidth}		\includegraphics[width=0.99\linewidth]{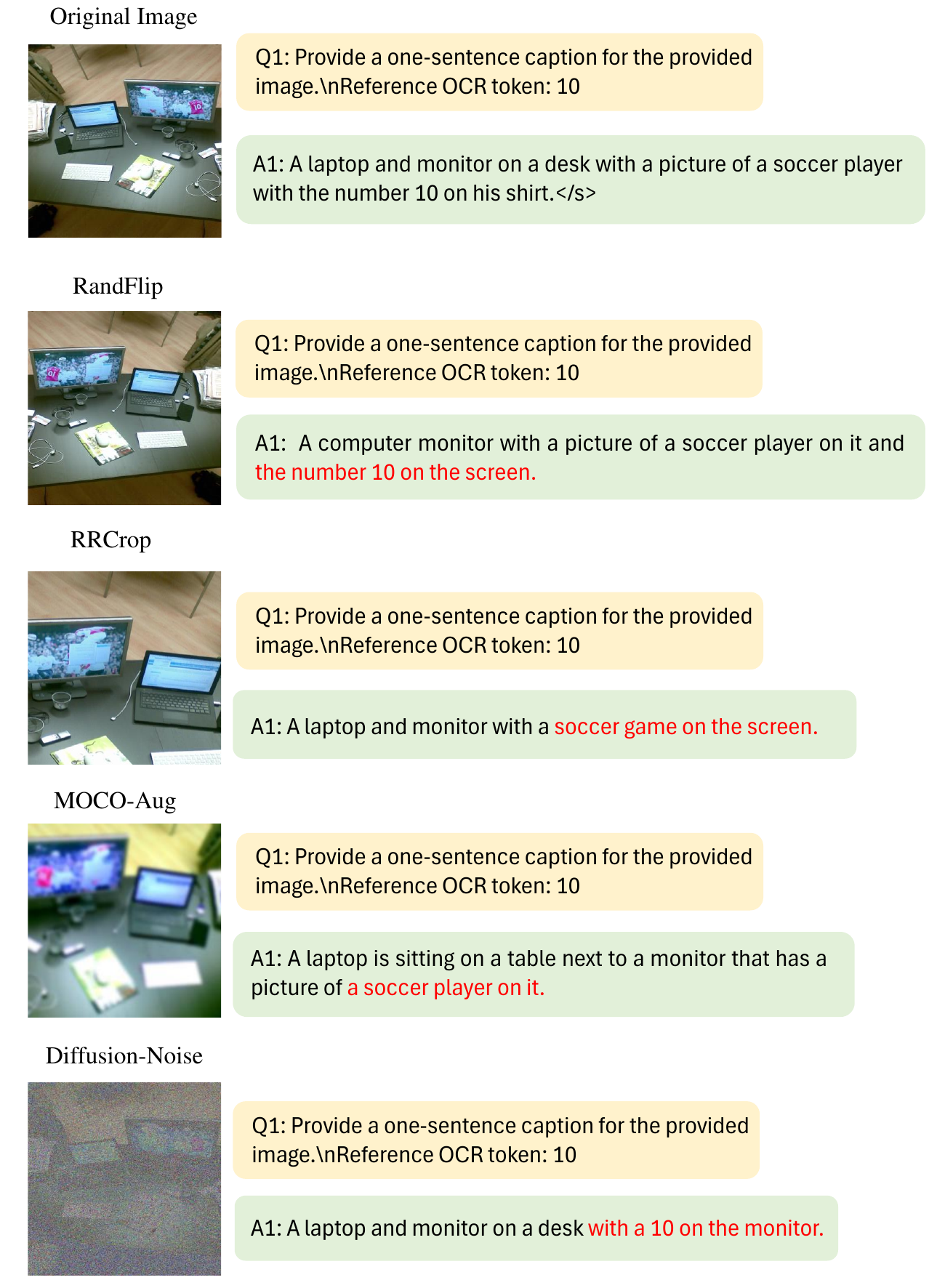}
    	\caption{data augmentations of TextVQA~\cite{TextVQA} images}
		\label{fig:}
    \end{subfigure} 
	\caption{Visualizations of the distorted images and their model answers. In-complete or in-correct tokens are highlighted with red colors.}
	\label{fig:augmentation-appendix}
\end{figure*}

\section{Data construction details} 
In Sec.~\ref{sec:exp-settings}, we have discussed the data construction process. Here, we want to emphasize the \emph{counting} of our data instance. Note that for clarity in all our experiment, we choose the num of the unfiltered preference sample as the DPO instances we used. But please note that the actual preference data sent to DPO are much less than that. One is the SeVa's filtering mechanism as discussed before (the filtering process in Alg.~\ref{alg:code} almost cut the instance num by half). The other is the data counting difference between LLaVA and SeVa: an instance of multi-turn dialogue will contribute to about 2 raw image-question pairs in our SeVa. As such, the data used in SeVa \emph{can be} far less counted if we `regard' this 2 image-question pairs as only one data instance that is sourced from a multi-dialogue.

\section{More visualizations.}

\subsection{Data augmentations samples}
In Fig.~\ref{fig:augmentation-appendix}, we visualize the data augmentation samples to show how the negatives responses are obtained. We choose original image, random-flip (`RandFlip'), random-resized-crop (`RRCrop'), moco augmentations and diffusion noise for better visualizations. We highlight those in-correct of in-complete answers with red color.

\subsection{GPT-4 prompt template}
Please refer to Fig.~\ref{fig:question-answer-appendix} for the GPT-4 prompt template we adopted to evaluate the consistency score in Table~\ref{tab:ablate-consistency}.

\subsection{More examples of SeVa}
\begin{figure*}
	\centering
    \includegraphics[width=0.8\linewidth]{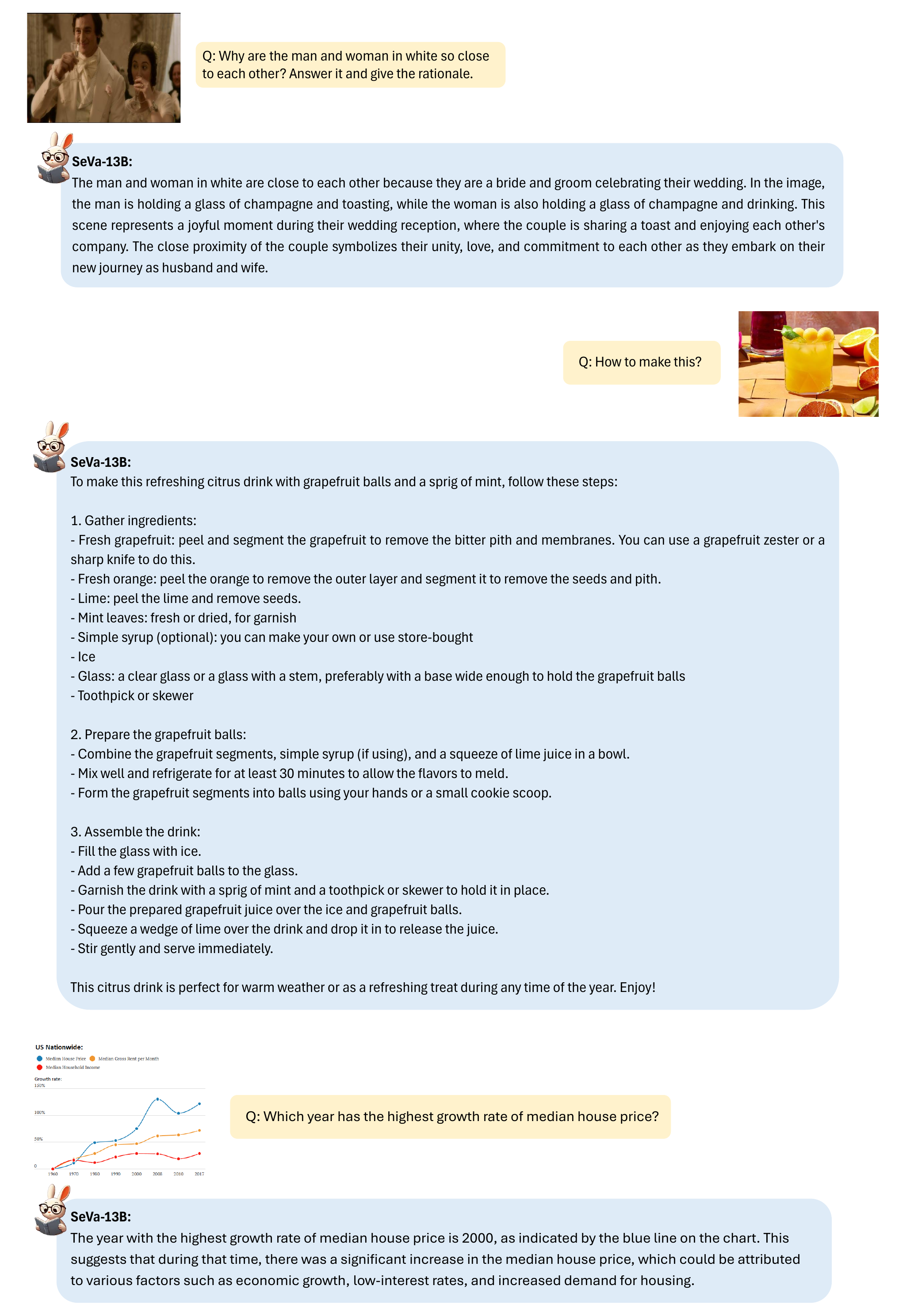}
	\caption{Visualizations of complex visual question answering with our SeVa-13B models.}
	\label{fig:vis-appendix}
\end{figure*}

\begin{figure*}
	\centering
    \includegraphics[width=0.8\linewidth]{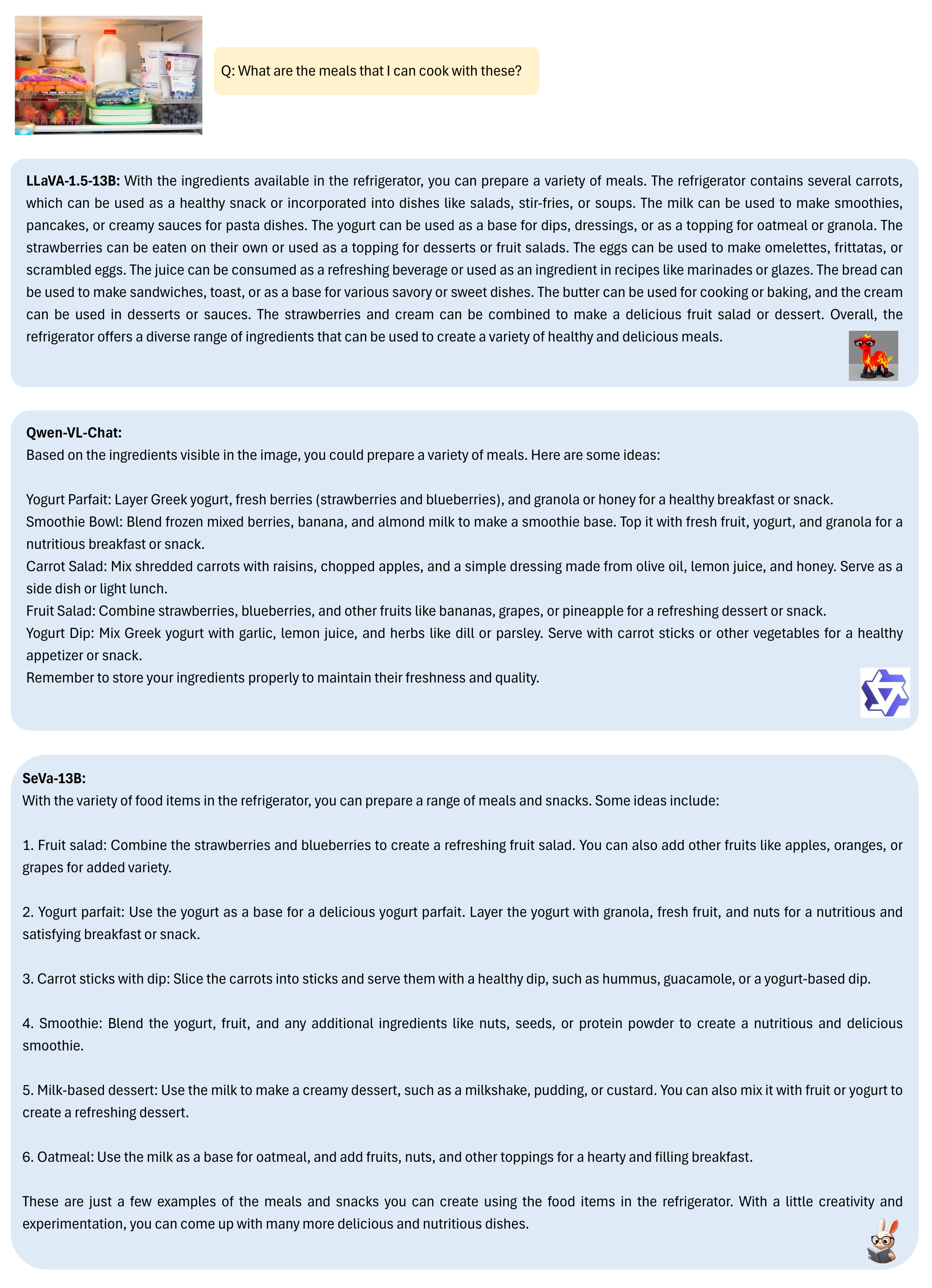}
	\caption{Comparison of our SeVa-13B with other VLMs (LLaVA-1.5, Qwen-VL-Chat).}
	\label{fig:vis-appendix1}
\end{figure*}

Please refer to Fig.~\ref{fig:vis-appendix}-\ref{fig:vis-appendix1}. In Fig.~\ref{fig:vis-appendix}, we showcase more examples of SeVa-13B to illustrate its multi-modal comprehension ability, including complex reasoning about a wedding, chain-of-thought power to make a drink and the capability to read scientific figures. In Fig.~\ref{fig:vis-appendix1}, we compare SeVa-13B with more VLMs, namely LLaVA-1.5-13B and Qwen-VL-chat. We found our model demonstrate a detailed and faithful answers, which is on par with (or even better than) Qwen-VL-chat that possibly requires more training data source and larger model capacity.


\clearpage
\clearpage
{
    \small
    \bibliographystyle{ieeenat_fullname}
    \bibliography{main}

\begin{thebibliography}{49}
\providecommand{\natexlab}[1]{#1}
\providecommand{\url}[1]{\texttt{#1}}
\expandafter\ifx\csname urlstyle\endcsname\relax
  \providecommand{\doi}[1]{doi: #1}\else
  \providecommand{\doi}{doi: \begingroup \urlstyle{rm}\Url}\fi

\bibitem[Adjali et~al.(2023)Adjali, Grimal, Ferret, Ghannay, and Le~Borgne]{ACM_VQA}
Omar Adjali, Paul Grimal, Olivier Ferret, Sahar Ghannay, and Herv{\'e} Le~Borgne.
\newblock Explicit knowledge integration for knowledge-aware visual question answering about named entities.
\newblock In \emph{Proceedings of the 2023 ACM International Conference on Multimedia Retrieval}, pages 29--38, 2023.

\bibitem[Alayrac et~al.(2022)Alayrac, Donahue, Luc, Miech, Barr, Hasson, Lenc, Mensch, Millican, Reynolds, Ring, Rutherford, Cabi, Han, Gong, Samangooei, Monteiro, Menick, Borgeaud, Brock, Nematzadeh, Sharifzadeh, Binkowski, Barreira, Vinyals, Zisserman, and Simonyan]{LLM_Flamingo}
Jean{-}Baptiste Alayrac, Jeff Donahue, Pauline Luc, Antoine Miech, Iain Barr, Yana Hasson, Karel Lenc, Arthur Mensch, Katherine Millican, Malcolm Reynolds, Roman Ring, Eliza Rutherford, Serkan Cabi, Tengda Han, Zhitao Gong, Sina Samangooei, Marianne Monteiro, Jacob~L. Menick, Sebastian Borgeaud, Andy Brock, Aida Nematzadeh, Sahand Sharifzadeh, Mikolaj Binkowski, Ricardo Barreira, Oriol Vinyals, Andrew Zisserman, and Kar{\'{e}}n Simonyan.
\newblock Flamingo: a visual language model for few-shot learning.
\newblock In \emph{Advances in Neural Information Processing Systems}, 2022.

\bibitem[Bai et~al.(2022)Bai, Kadavath, Kundu, Askell, Kernion, Jones, Chen, Goldie, Mirhoseini, McKinnon, et~al.]{RLAIF}
Yuntao Bai, Saurav Kadavath, Sandipan Kundu, Amanda Askell, Jackson Kernion, Andy Jones, Anna Chen, Anna Goldie, Azalia Mirhoseini, Cameron McKinnon, et~al.
\newblock Constitutional ai: Harmlessness from ai feedback.
\newblock \emph{arXiv preprint arXiv:2212.08073}, 2022.

\bibitem[Chen et~al.(2023)Chen, Li, Dong, Zhang, He, Wang, Zhao, and Lin]{ShareGPT4V}
Lin Chen, Jisong Li, Xiaoyi Dong, Pan Zhang, Conghui He, Jiaqi Wang, Feng Zhao, and Dahua Lin.
\newblock Sharegpt4v: Improving large multi-modal models with better captions.
\newblock \emph{arXiv preprint arXiv:2311.12793}, 2023.

\bibitem[Chen et~al.(2020)Chen, Kornblith, Norouzi, and Hinton]{SimCLR}
Ting Chen, Simon Kornblith, Mohammad Norouzi, and Geoffrey Hinton.
\newblock A simple framework for contrastive learning of visual representations.
\newblock pages 1597--1607, 2020.

\bibitem[Chiang et~al.(2023)Chiang, Li, Lin, Sheng, Wu, Zhang, Zheng, Zhuang, Zhuang, Gonzalez, Stoica, and Xing]{vicuna2023}
Wei-Lin Chiang, Zhuohan Li, Zi Lin, Ying Sheng, Zhanghao Wu, Hao Zhang, Lianmin Zheng, Siyuan Zhuang, Yonghao Zhuang, Joseph~E. Gonzalez, Ion Stoica, and Eric~P. Xing.
\newblock Vicuna: An open-source chatbot impressing gpt-4 with 90\%* chatgpt quality, 2023.

\bibitem[Contributors(2023)]{VLMevalkit}
OpenCompass Contributors.
\newblock Opencompass: A universal evaluation platform for foundation models.
\newblock \url{https://github.com/open-compass/opencompass}, 2023.

\bibitem[Dai et~al.(2023)Dai, Li, LI, Tiong, Zhao, Wang, Li, Fung, and Hoi]{InstructBLIP}
Wenliang Dai, Junnan Li, DONGXU LI, Anthony Tiong, Junqi Zhao, Weisheng Wang, Boyang Li, Pascale~N Fung, and Steven Hoi.
\newblock Instructblip: Towards general-purpose vision-language models with instruction tuning.
\newblock In \emph{Advances in Neural Information Processing Systems}, pages 49250--49267, 2023.

\bibitem[Dong et~al.(2024)Dong, Han, Peng, Qi, Ge, Yang, Zhao, Sun, Zhou, Wei, Kong, Zhang, Ma, and Yi]{dreamllm}
Runpei Dong, Chunrui Han, Yuang Peng, Zekun Qi, Zheng Ge, Jinrong Yang, Liang Zhao, Jianjian Sun, Hongyu Zhou, Haoran Wei, Xiangwen Kong, Xiangyu Zhang, Kaisheng Ma, and Li Yi.
\newblock Dream{LLM}: Synergistic multimodal comprehension and creation.
\newblock In \emph{The Twelfth International Conference on Learning Representations}, 2024.

\bibitem[Fu et~al.(2024)Fu, Zhu, and Wu]{DTL}
Minghao Fu, Ke Zhu, and Jianxin Wu.
\newblock Dtl: Disentangled transfer learning for visual recognition.
\newblock In \emph{Proceedings of the AAAI Conference on Artificial Intelligence}, pages 12082--12090, 2024.

\bibitem[Ge et~al.(2023)Ge, Zhou, Hou, Khabsa, Wang, Wang, Han, and Mao]{redteaming}
Suyu Ge, Chunting Zhou, Rui Hou, Madian Khabsa, Yi-Chia Wang, Qifan Wang, Jiawei Han, and Yuning Mao.
\newblock Mart: Improving llm safety with multi-round automatic red-teaming.
\newblock \emph{arXiv preprint arXiv:2311.07689}, 2023.

\bibitem[Grill et~al.(2020)Grill, Strub, Altch{\'e}, Tallec, Richemond, Buchatskaya, Doersch, Avila~Pires, Guo, Gheshlaghi~Azar, et~al.]{BYOL}
Jean-Bastien Grill, Florian Strub, Florent Altch{\'e}, Corentin Tallec, Pierre Richemond, Elena Buchatskaya, Carl Doersch, Bernardo Avila~Pires, Zhaohan Guo, Mohammad Gheshlaghi~Azar, et~al.
\newblock Bootstrap your own latent-a new approach to self-supervised learning.
\newblock In \emph{NeurIPS}, pages 21271--21284, 2020.

\bibitem[Gulcehre et~al.(2023)Gulcehre, Paine, Srinivasan, Konyushkova, Weerts, Sharma, Siddhant, Ahern, Wang, Gu, et~al.]{ReST}
Caglar Gulcehre, Tom~Le Paine, Srivatsan Srinivasan, Ksenia Konyushkova, Lotte Weerts, Abhishek Sharma, Aditya Siddhant, Alex Ahern, Miaosen Wang, Chenjie Gu, et~al.
\newblock Reinforced self-training (rest) for language modeling.
\newblock \emph{arXiv preprint arXiv:2308.08998}, 2023.

\bibitem[He et~al.(2020)He, Fan, Wu, Xie, and Girshick]{MOCO}
Kaiming He, Haoqi Fan, Yuxin Wu, Saining Xie, and Ross Girshick.
\newblock Momentum contrast for unsupervised visual representation learning.
\newblock In \emph{CVPR}, pages 9729--9738, 2020.

\bibitem[Hu et~al.(2021)Hu, Wallis, Allen-Zhu, Li, Wang, Wang, Chen, et~al.]{LoRA}
Edward~J Hu, Phillip Wallis, Zeyuan Allen-Zhu, Yuanzhi Li, Shean Wang, Lu Wang, Weizhu Chen, et~al.
\newblock Lora: Low-rank adaptation of large language models.
\newblock In \emph{International Conference on Learning Representations}, 2021.

\bibitem[Hudson and Manning(2019)]{GQA}
Drew~A Hudson and Christopher~D Manning.
\newblock {GQA}: A new dataset for real-world visual reasoning and compositional question answering.
\newblock In \emph{Proceedings of the IEEE/CVF conference on computer vision and pattern recognition}, pages 6700--6709, 2019.

\bibitem[Leng et~al.(2023)Leng, Zhang, Chen, Li, Lu, Miao, and Bing]{VCD}
Sicong Leng, Hang Zhang, Guanzheng Chen, Xin Li, Shijian Lu, Chunyan Miao, and Lidong Bing.
\newblock Mitigating object hallucinations in large vision-language models through visual contrastive decoding.
\newblock \emph{arXiv preprint arXiv:2311.16922}, 2023.

\bibitem[Li et~al.(2023{\natexlab{a}})Li, Wang, Wang, Ge, Ge, and Shan]{Seed-Bench}
Bohao Li, Rui Wang, Guangzhi Wang, Yuying Ge, Yixiao Ge, and Ying Shan.
\newblock Seed-bench: Benchmarking multimodal llms with generative comprehension.
\newblock \emph{arXiv preprint arXiv:2307.16125}, 2023{\natexlab{a}}.

\bibitem[Li et~al.(2024)Li, Lin, and Pei]{VLLM_DPO}
Shengzhi Li, Rongyu Lin, and Shichao Pei.
\newblock Multi-modal preference alignment remedies regression of visual instruction tuning on language model.
\newblock \emph{arXiv preprint arXiv:2402.10884}, 2024.

\bibitem[Li et~al.(2023{\natexlab{b}})Li, Du, Zhou, Wang, Zhao, and Wen]{POPE}
Yifan Li, Yifan Du, Kun Zhou, Jinpeng Wang, Wayne~Xin Zhao, and Ji-Rong Wen.
\newblock Evaluating object hallucination in large vision-language models.
\newblock In \emph{Proceedings of the 2023 Conference on Empirical Methods in Natural Language Processing}, pages 292--305, 2023{\natexlab{b}}.

\bibitem[Li et~al.(2023{\natexlab{c}})Li, Wang, and Jia]{LLaMA-VID}
Yanwei Li, Chengyao Wang, and Jiaya Jia.
\newblock Llama-vid: An image is worth 2 tokens in large language models.
\newblock \emph{arXiv preprint arXiv:2311.17043}, 2023{\natexlab{c}}.

\bibitem[Li et~al.(2023{\natexlab{d}})Li, Zhang, Yu, Wang, Fu, Lin, Shen, Chen, and Wei]{DiffusionSFT}
Yanda Li, Chi Zhang, Gang Yu, Zhibin Wang, Bin Fu, Guosheng Lin, Chunhua Shen, Ling Chen, and Yunchao Wei.
\newblock Stablellava: Enhanced visual instruction tuning with synthesized image-dialogue data.
\newblock \emph{arXiv preprint arXiv:2308.10253}, 2023{\natexlab{d}}.

\bibitem[Lin et~al.(2023)Lin, Yin, Ping, Lu, Molchanov, Tao, Mao, Kautz, Shoeybi, and Han]{VILA}
Ji Lin, Hongxu Yin, Wei Ping, Yao Lu, Pavlo Molchanov, Andrew Tao, Huizi Mao, Jan Kautz, Mohammad Shoeybi, and Song Han.
\newblock Vila: On pre-training for visual language models.
\newblock \emph{arXiv preprint arXiv:2312.07533}, 2023.

\bibitem[Liu et~al.(2023{\natexlab{a}})Liu, Li, Li, and Lee]{LLaVa1.5}
Haotian Liu, Chunyuan Li, Yuheng Li, and Yong~Jae Lee.
\newblock Improved baselines with visual instruction tuning.
\newblock \emph{arXiv preprint arXiv:2310.03744}, 2023{\natexlab{a}}.

\bibitem[Liu et~al.(2024)Liu, Li, Wu, and Lee]{LLM_Llava}
Haotian Liu, Chunyuan Li, Qingyang Wu, and Yong~Jae Lee.
\newblock Visual instruction tuning.
\newblock \emph{Advances in neural information processing systems}, 36, 2024.

\bibitem[Liu et~al.(2023{\natexlab{b}})Liu, Duan, Zhang, Li, Zhang, Zhao, Yuan, Wang, He, Liu, et~al.]{MMBench}
Yuan Liu, Haodong Duan, Yuanhan Zhang, Bo Li, Songyang Zhang, Wangbo Zhao, Yike Yuan, Jiaqi Wang, Conghui He, Ziwei Liu, et~al.
\newblock Mmbench: Is your multi-modal model an all-around player?
\newblock \emph{arXiv preprint arXiv:2307.06281}, 2023{\natexlab{b}}.

\bibitem[Lu et~al.(2022)Lu, Mishra, Xia, Qiu, Chang, Zhu, Tafjord, Clark, and Kalyan]{SQA}
Pan Lu, Swaroop Mishra, Tony Xia, Liang Qiu, Kai-Wei Chang, Song-Chun Zhu, Oyvind Tafjord, Peter Clark, and Ashwin Kalyan.
\newblock Learn to explain: Multimodal reasoning via thought chains for science question answering.
\newblock In \emph{The 36th Conference on Neural Information Processing Systems (NeurIPS)}, 2022.

\bibitem[Mishra et~al.(2019)Mishra, Shekhar, Singh, and Chakraborty]{OCRVQA}
Anand Mishra, Shashank Shekhar, Ajeet~Kumar Singh, and Anirban Chakraborty.
\newblock Ocr-vqa: Visual question answering by reading text in images.
\newblock In \emph{2019 international conference on document analysis and recognition (ICDAR)}, pages 947--952. IEEE, 2019.

\bibitem[Ouyang et~al.(2022)Ouyang, Wu, Jiang, Almeida, Wainwright, Mishkin, Zhang, Agarwal, Slama, Ray, et~al.]{InstructGPT}
Long Ouyang, Jeffrey Wu, Xu Jiang, Diogo Almeida, Carroll Wainwright, Pamela Mishkin, Chong Zhang, Sandhini Agarwal, Katarina Slama, Alex Ray, et~al.
\newblock Training language models to follow instructions with human feedback.
\newblock \emph{Advances in neural information processing systems}, 35:\penalty0 27730--27744, 2022.

\bibitem[Radford et~al.(2021)Radford, Kim, Hallacy, Ramesh, Goh, Agarwal, Sastry, Askell, Mishkin, Clark, et~al.]{CLIP}
Alec Radford, Jong~Wook Kim, Chris Hallacy, Aditya Ramesh, Gabriel Goh, Sandhini Agarwal, Girish Sastry, Amanda Askell, Pamela Mishkin, Jack Clark, et~al.
\newblock Learning transferable visual models from natural language supervision.
\newblock In \emph{International conference on machine learning}, pages 8748--8763. PMLR, 2021.

\bibitem[Rafailov et~al.(2024)Rafailov, Sharma, Mitchell, Manning, Ermon, and Finn]{DPO}
Rafael Rafailov, Archit Sharma, Eric Mitchell, Christopher~D Manning, Stefano Ermon, and Chelsea Finn.
\newblock Direct preference optimization: Your language model is secretly a reward model.
\newblock 2024.

\bibitem[Schulman et~al.(2017)Schulman, Wolski, Dhariwal, Radford, and Klimov]{PPO}
John Schulman, Filip Wolski, Prafulla Dhariwal, Alec Radford, and Oleg Klimov.
\newblock Proximal policy optimization algorithms.
\newblock \emph{arXiv preprint arXiv:1707.06347}, 2017.

\bibitem[Singh et~al.(2019)Singh, Natarajan, Shah, Jiang, Chen, Batra, Parikh, and Rohrbach]{TextVQA}
Amanpreet Singh, Vivek Natarajan, Meet Shah, Yu Jiang, Xinlei Chen, Dhruv Batra, Devi Parikh, and Marcus Rohrbach.
\newblock Towards vqa models that can read.
\newblock In \emph{Proceedings of the IEEE/CVF conference on computer vision and pattern recognition}, pages 8317--8326, 2019.

\bibitem[Sun et~al.(2023{\natexlab{a}})Sun, Li, and Dong]{ACM_Retrieval}
Lina Sun, Yewen Li, and Yumin Dong.
\newblock Learning from expert: Vision-language knowledge distillation for unsupervised cross-modal hashing retrieval.
\newblock In \emph{Proceedings of the 2023 ACM International Conference on Multimedia Retrieval}, pages 499--507, 2023{\natexlab{a}}.

\bibitem[Sun et~al.(2023{\natexlab{b}})Sun, Shen, Cao, Liu, Li, Shen, Gan, Gui, Wang, Yang, et~al.]{LLaVa-RLHF}
Zhiqing Sun, Sheng Shen, Shengcao Cao, Haotian Liu, Chunyuan Li, Yikang Shen, Chuang Gan, Liang-Yan Gui, Yu-Xiong Wang, Yiming Yang, et~al.
\newblock Aligning large multimodal models with factually augmented rlhf.
\newblock \emph{arXiv preprint arXiv:2309.14525}, 2023{\natexlab{b}}.

\bibitem[Wang et~al.(2023)Wang, Meng, Weng, He, Wu, and Jiang]{LVIS-Instruct4V}
Junke Wang, Lingchen Meng, Zejia Weng, Bo He, Zuxuan Wu, and Yu-Gang Jiang.
\newblock To see is to believe: Prompting gpt-4v for better visual instruction tuning.
\newblock \emph{arXiv preprint arXiv:2311.07574}, 2023.

\bibitem[Wei et~al.(2023)Wei, Kong, Chen, Zhao, Ge, Yang, Sun, Han, and Zhang]{vary}
Haoran Wei, Lingyu Kong, Jinyue Chen, Liang Zhao, Zheng Ge, Jinrong Yang, Jianjian Sun, Chunrui Han, and Xiangyu Zhang.
\newblock Vary: Scaling up the vision vocabulary for large vision-language models.
\newblock \emph{arXiv preprint arXiv:2312.06109}, 2023.

\bibitem[Yu et~al.(2023{\natexlab{a}})Yu, Zhao, Wei, Yang, Wu, Kong, Wei, Wang, Ge, Zhang, et~al.]{merlin}
En Yu, Liang Zhao, Yana Wei, Jinrong Yang, Dongming Wu, Lingyu Kong, Haoran Wei, Tiancai Wang, Zheng Ge, Xiangyu Zhang, et~al.
\newblock Merlin: Empowering multimodal llms with foresight minds.
\newblock \emph{arXiv preprint arXiv:2312.00589}, 2023{\natexlab{a}}.

\bibitem[Yu et~al.(2023{\natexlab{b}})Yu, Yang, Li, Wang, Lin, Liu, Wang, and Wang]{MMVet}
Weihao Yu, Zhengyuan Yang, Linjie Li, Jianfeng Wang, Kevin Lin, Zicheng Liu, Xinchao Wang, and Lijuan Wang.
\newblock Mm-vet: Evaluating large multimodal models for integrated capabilities.
\newblock \emph{arXiv preprint arXiv:2308.02490}, 2023{\natexlab{b}}.

\bibitem[Zachariah and Rao(2023)]{ACM_VideoRetrieval}
Arun Zachariah and Praveen Rao.
\newblock Video retrieval for everyday scenes with common objects.
\newblock In \emph{Proceedings of the 2023 ACM International Conference on Multimedia Retrieval}, pages 565--570, 2023.

\bibitem[Zhang et~al.(2024)Zhang, Gui, Sun, Feng, Xu, Zhang, Fu, Li, Hauptmann, Bisk, et~al.]{zhaoAlign}
Ruohong Zhang, Liangke Gui, Zhiqing Sun, Yihao Feng, Keyang Xu, Yuanhan Zhang, Di Fu, Chunyuan Li, Alexander Hauptmann, Yonatan Bisk, et~al.
\newblock Direct preference optimization of video large multimodal models from language model reward.
\newblock \emph{arXiv preprint arXiv:2404.01258}, 2024.

\bibitem[Zhao et~al.(2023{\natexlab{a}})Zhao, Wu, and Huang]{Scale_SFT}
Bo Zhao, Boya Wu, and Tiejun Huang.
\newblock Svit: Scaling up visual instruction tuning.
\newblock \emph{arXiv preprint arXiv:2307.04087}, 2023{\natexlab{a}}.

\bibitem[Zhao et~al.(2023{\natexlab{b}})Zhao, Yu, Ge, Yang, Wei, Zhou, Sun, Peng, Dong, Han, et~al.]{chatspot}
Liang Zhao, En Yu, Zheng Ge, Jinrong Yang, Haoran Wei, Hongyu Zhou, Jianjian Sun, Yuang Peng, Runpei Dong, Chunrui Han, et~al.
\newblock Chatspot: Bootstrapping multimodal llms via precise referring instruction tuning.
\newblock \emph{arXiv preprint arXiv:2307.09474}, 2023{\natexlab{b}}.

\bibitem[Zhao et~al.(2023{\natexlab{c}})Zhao, Wang, Ouyang, Dong, Wang, and He]{HA-DPO}
Zhiyuan Zhao, Bin Wang, Linke Ouyang, Xiaoyi Dong, Jiaqi Wang, and Conghui He.
\newblock Beyond hallucinations: Enhancing lvlms through hallucination-aware direct preference optimization.
\newblock \emph{arXiv preprint arXiv:2311.16839}, 2023{\natexlab{c}}.

\bibitem[Zheng et~al.(2024)Zheng, Chiang, Sheng, Zhuang, Wu, Zhuang, Lin, Li, Li, Xing, et~al.]{LLM-as-judge}
Lianmin Zheng, Wei-Lin Chiang, Ying Sheng, Siyuan Zhuang, Zhanghao Wu, Yonghao Zhuang, Zi Lin, Zhuohan Li, Dacheng Li, Eric Xing, et~al.
\newblock Judging llm-as-a-judge with mt-bench and chatbot arena.
\newblock \emph{Advances in Neural Information Processing Systems}, 36, 2024.

\bibitem[Zhu et~al.(2023{\natexlab{a}})Zhu, Chen, Shen, Li, and Elhoseiny]{LLM_MiniGPT4}
Deyao Zhu, Jun Chen, Xiaoqian Shen, Xiang Li, and Mohamed Elhoseiny.
\newblock Minigpt-4: Enhancing vision-language understanding with advanced large language models.
\newblock \emph{arXiv preprint arXiv:2304.10592}, 2023{\natexlab{a}}.

\bibitem[Zhu et~al.(2023{\natexlab{b}})Zhu, Fu, and Wu]{MLS}
Ke Zhu, Minghao Fu, and Jianxin Wu.
\newblock Multi-label self-supervised learning with scene images.
\newblock In \emph{Proceedings of the IEEE/CVF International Conference on Computer Vision}, pages 6694--6703, 2023{\natexlab{b}}.

\bibitem[Zhu et~al.(2023{\natexlab{c}})Zhu, He, and Wu]{Cropping}
Ke Zhu, Yin-Yin He, and Jianxin Wu.
\newblock Coarse is better? a new pipeline towards self-supervised learning with uncurated images.
\newblock \emph{arXiv preprint arXiv:2306.04244}, 2023{\natexlab{c}}.

\bibitem[Zong et~al.(2024)Zong, Bohdal, Yu, Yang, and Hospedales]{SafelyFT}
Yongshuo Zong, Ondrej Bohdal, Tingyang Yu, Yongxin Yang, and Timothy Hospedales.
\newblock Safety fine-tuning at (almost) no cost: A baseline for vision large language models.
\newblock \emph{arXiv preprint arXiv:2402.02207}, 2024.

\end{thebibliography}
}


\end{document}